\documentclass[10pt,twocolumn,letterpaper]{article}

\usepackage{arxiv}
\newtoggle{usetimes}
\newtoggle{itabs}
\toggletrue{usetimes}  

\iftoggle{usetimes}{\usepackage{times}}{}
\usepackage{amsmath,amssymb,amsfonts,dsfont,pifont,bm,bbm,mathrsfs,mathtools,nicefrac}
\usepackage{algorithm,listings}
\usepackage[noend]{algpseudocode}
\usepackage{booktabs,multirow,adjustbox,diagbox,threeparttable,makecell}
\definecolor{citeblue}{RGB}{48,111,186}
\usepackage[pagebackref=false,breaklinks=true,colorlinks=true,citecolor=citeblue,bookmarks=false]{hyperref}
\usepackage{cleveref}  

\usepackage{amsmath,amsfonts,bm}
\def\T{T}
\def\defeq{\triangleq}

\def\vxi{{\bm{\xi}}}

\def\dxkt#1#2{{\rvy}_{{#1},{#2}}}

\def\eqref#1{equation~\ref{#1}}

\def\1{\bm{1}}

\def\rvepsilon{{\mathbf{\epsilon}}}

\def\rvv{{\mathbf{v}}}

\def\rvx{{\mathbf{x}}}
\def\rvy{{\mathbf{y}}}
\def\rvz{{\mathbf{z}}}

\def\vzero{{\bm{0}}}

\def\vmu{{\bm{\mu}}}

\def\veps{{\bm{\epsilon}}}

\def\vx{{\bm{x}}}

\def\mA{{\bm{A}}}
\def\mB{{\bm{B}}}

\def\mG{{\bm{G}}}

\def\mI{{\bm{I}}}

\def\mL{{\bm{L}}}

\def\mN{{\bm{N}}}
\def\mO{{\bm{O}}}

\def\mU{{\bm{U}}}

\def\mW{{\bm{W}}}

\def\mLambda{{\bm{\Lambda}}}
\def\mSigma{{\bm{\Sigma}}}

\DeclareMathAlphabet{\mathsfit}{\encodingdefault}{\sfdefault}{m}{sl}
\SetMathAlphabet{\mathsfit}{bold}{\encodingdefault}{\sfdefault}{bx}{n}

\def\gD{{\mathcal{D}}}

\def\gN{{\mathcal{N}}}

\def\gU{{\mathcal{U}}}

\def\sD{{\mathbb{D}}}

\def\sN{{\mathbb{N}}}

\def\sR{{\mathbb{R}}}
\def\sS{{\mathbb{S}}}
\def\sT{{\mathbb{T}}}

\newcommand{\E}{\mathbb{E}}

\newtheorem{theorem}{Theorem}
\newtheorem{proposition}{Proposition}
\crefname{section}{Sec.}{Secs.}
\Crefname{section}{Section}{Sections}
\crefname{table}{Tab.}{Tabs.}
\Crefname{table}{Table}{Tables}
\crefname{figure}{Fig.}{Figs.}
\Crefname{figure}{Figure}{Figures}
\crefname{equation}{Eq.}{Eqs.}
\Crefname{equation}{Equation}{Equations}
\crefname{theorem}{Thm.}{Thms.}
\Crefname{theorem}{Theorem}{Theorems}
\crefname{algorithm}{Alg.}{Algs.}
\Crefname{algorithm}{Algorithm}{Algorithms}
\newcommand{\tocite}[1]{\textcolor{red}{[TO CITE]}}
\newcommand{\method}{dimensionality-varying diffusion process\xspace}
\newcommand{\Method}{Dimensionality-Varying Diffusion Process\xspace}
\newcommand{\methodabbr}{DVDP\xspace}

\newcommand{\Adp}{Attenuated Diffusion Process\xspace}
\newcommand{\adpabbr}{ADP\xspace}

\newcommand\nonumfootnote[1]{%
\begingroup%
    \renewcommand\thefootnote{}\footnote{\hspace{-4pt}#1}%
    \addtocounter{footnote}{-1}%
\endgroup%
}

\usepackage{listings}
\definecolor{codegreen}{rgb}{0,0.6,0}
\definecolor{codegray}{rgb}{0.5,0.5,0.5}
\definecolor{codepurple}{rgb}{0.58,0,0.82}
\definecolor{backcolour}{rgb}{1.0,1.0,1.0}
\lstdefinestyle{mystyle}{
    backgroundcolor=\color{backcolour},
    commentstyle=\color{codegreen},
    keywordstyle=\color{magenta},
    numberstyle=\tiny\color{codegray},
    stringstyle=\color{codepurple},
    basicstyle=\ttfamily\scriptsize,
    breakatwhitespace=false,
    breaklines=true,
    captionpos=b,
    keepspaces=true,
    numbers=left,
    numbersep=5pt,
    showspaces=false,
    showstringspaces=false,
    showtabs=false,
    tabsize=2
}
\begin{document}

\title{\Method}

\author{
    Han Zhang$^{1,3}$ \quad
    Ruili Feng$^{2,3}$ \quad
    Zhantao Yang$^{1,3}$ \quad
    Lianghua Huang$^3$ \quad
    Yu Liu$^3$
    \\[0pt]
    Yifei Zhang$^{1,3}$ \quad
    Yujun Shen$^4$ \quad
    Deli Zhao$^3$ \quad
    Jingren Zhou$^3$ \quad
    Fan Cheng$^{1*}$
    \\[5pt]
    $^1$SJTU \quad 
    $^2$USTC \quad 
    $^3$Alibaba Group \quad 
    $^4$Ant Group
    \\[5pt]
    \footnotesize\texttt{\{hzhang9617, ruilifengustc, ztyang196\}@gmail.com} \quad
    \footnotesize\texttt{\{xuangen.hlh, ly103369\}@alibaba-inc.com}
    \\[0pt]
    \footnotesize\texttt{qidouxiong619@sjtu.edu.cn} \quad
    \footnotesize\texttt{\{shenyujun0302, zhaodeli\}@gmail.com}
    \\[0pt]
    \footnotesize\texttt{jingren.zhou@alibaba-inc.com} \quad
    \footnotesize\texttt{chengfan@sjtu.edu.cn}
}

\onecolumn

\maketitle

\begin{figure}[!ht]
	\centering
	\vspace{-10pt}
	\includegraphics[width=1.0\linewidth]{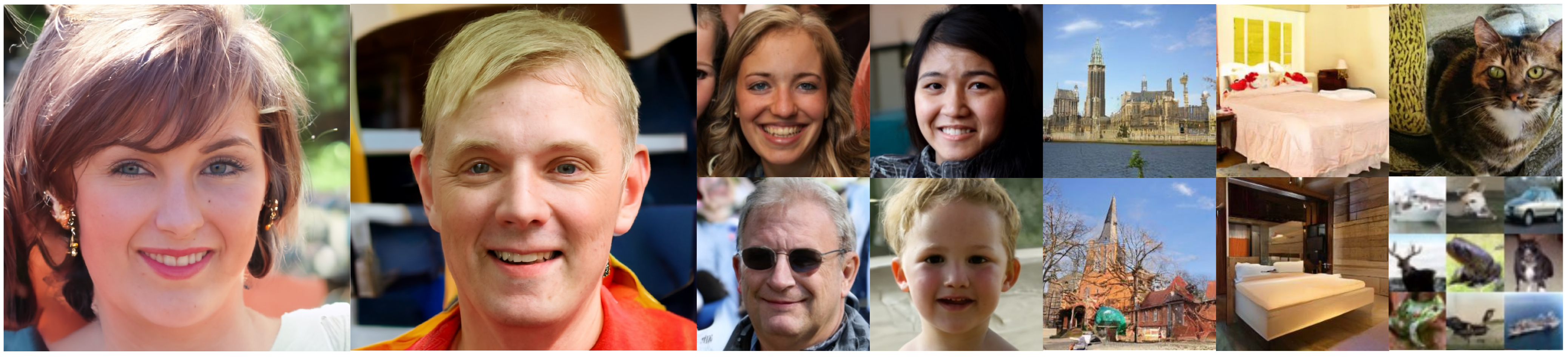}
	\vspace{-20pt}
        \caption{
            Synthesized samples on various datasets, including FFHQ ($1024^2$ and $256^2$), LSUN Church ($256^2$), LSUN Bedroom ($256^2$), LSUN Cat 256 ($256^2$) and CIFAR10 ($32^2$). All these samples are generated from a $64^2$ noise except CIFAR10 from $16^2$, while conventional diffusion models can only start from a noise with the same dimension as the final sample.
        }
        \label{figure:generative-samples}
	\vspace{10pt}
\end{figure}

\begin{abstract}

\iftoggle{itabs}{\it}{}

Diffusion models, which learn to reverse a signal destruction process to generate new data, typically require the signal at each step to have the same dimension.
We argue that, considering the spatial redundancy in image signals, there is no need to maintain a high dimensionality in the evolution process, especially in the early generation phase.
To this end, we make a theoretical generalization of the forward diffusion process via signal decomposition.
Concretely, we manage to decompose an image into multiple orthogonal components and control the attenuation of each component when perturbing the image.
That way, along with the noise strength increasing, we are able to diminish those inconsequential components and thus use a lower-dimensional signal to represent the source, barely losing information.
Such a reformulation allows to vary dimensions in both training and inference of diffusion models.
Extensive experiments on a range of datasets suggest that our approach substantially reduces the computational cost and achieves on-par or even better synthesis performance compared to baseline methods.
We also show that our strategy facilitates high-resolution image synthesis and improves FID of diffusion model trained on FFHQ at $1024\times1024$ resolution from 52.40 to 10.46.
Code and models will be made publicly available.
%
\nonumfootnote{$^*$corresponding author}
%

\end{abstract}

\section{Introduction}\label{sec:intro}

\begin{figure*}[t]
\begin{center}
\includegraphics[width=0.8\textwidth]{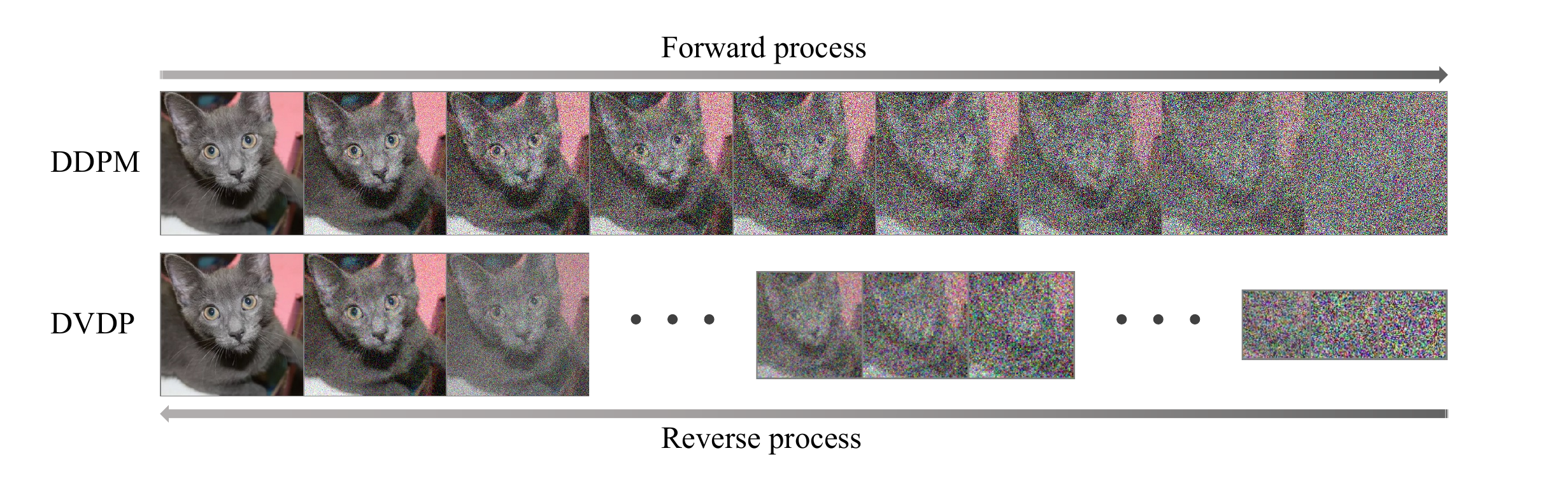}
\end{center}
\vspace{-15pt}
\caption{
    \textbf{Conceptual comparison} between DDPM~\cite{ho2020denoising} and our proposed \methodabbr, where our approach allows using a varying dimension in the diffusion process.
}
\label{fig:generative-process-comparison}
\vspace{-5pt}
\end{figure*}

Diffusion models~\cite{ho2020denoising, song2020score, dhariwal2021diffusion, ramesh2022hierarchical, saharia2022photorealistic} have recently shown great potential in image synthesis.
Instead of directly learning the observed distribution, it constructs a multi-step forward process through gradually adding noise onto the real data (\textit{i.e.}, diffusion).
After a sufficiently large number of steps, the source signal could be considered as completely destroyed, resulting in a pure noise distribution that naturally supports sampling.
In this way, starting from sampled noises, we can expect new instances after reversing the diffusion process step by step.

As it can be seen, the above pipeline does not change the dimension of the source signal throughout the entire diffusion process~\cite{sohl2015deep, ho2020denoising, song2020score}.
It thus requires the reverse process to map a high-dimensional input to a high-dimensional output at every single step, causing heavy computation overheads~\cite{rombach2022high, jing2022subspace}.
However, images present a measure of spatial redundancy~\cite{he2022masked} from the semantic perspective (\textit{e.g.}, an image pixel could usually be easily predicted according to its neighbours).
Given such a fact, when the source signal is attenuated to some extent along with the noise strength growing, it should be possible to get replaced by a lower-dimensional signal.
We therefore argue that there is no need to follow the source signal dimension along the entire distribution evolution process, especially at early steps (\textit{i.e.}, steps close to the pure noise distribution) for coarse generation.

In this work, we propose \method (\methodabbr), which allows dynamically adjusting the signal dimension when constructing the forward path.
For this purpose, we first decompose an image into multiple orthogonal components, each of which owns  dimension lower than the original data.
Then, based on such a decomposition, we theoretically generalize the conventional diffusion process such that we can control the attenuation of each component when adding noise.
Thanks to this reformulation, we manage to drop those inconsequential components after the noise strength reaches a certain level, and thus represent the source image using a lower-dimensional signal with little information lost.
The remaining diffusion process could inherit this dimension and apply the same technique to further reduce the dimension.

We evaluate our approach on various datasets, including objects, human faces, animals, indoor scenes, and outdoor scenes.
Experimental results suggest that \methodabbr achieves on-par or even better synthesis performance than baseline models on all datasets.
More importantly, \methodabbr relies on much fewer computations, and hence speeds up both training and inference of diffusion models.
We also demonstrate the effectiveness of our approach in learning from high-resolution data.
For example, we are able to start from a $64\times64$ noise to produce an image under $1024\times1024$ resolution.
With FID~\cite{heusel2017gans} as the evaluation metric, our $1024\times1024$ model trained on FFHQ improves the baseline~\cite{song2020score} from 52.40 to 10.46.
All these advantages benefit from using a lower-dimensional signal, which reduces the computational cost and mitigates the optimization difficulty.

\section{Related Work}\label{sec:related-work}

\noindent\textbf{Diffusion models.}
Sohl-Dickstein et al.~\cite{sohl2015deep} propose diffusion models for the first time that generate samples from a target distribution by reversing a diffusion process in which target distribution is gradually disturbed to an easily sampled standard Gaussian. Ho et al.~\cite{ho2020denoising} further propose DDPM to reverse the diffusion process by learning a noise prediction network. Song et al.~\cite{song2020score} consider diffusion models as stochastic differential equations with continuous timesteps and proposes a unified framework. 

\vspace{2pt}
\noindent\textbf{Accelerating diffusion models.}
Diffusion models significantly suffer from the slow training and inference speed. There are many methods that speed up sampling from thousands of steps to tens of steps while keeping an acceptable sample quality~\cite{bao2022analytic, lu2022dpm, song2020denoising, nichol2021improved, watson2021learning, watson2021learning1, san2021noise, liu2022pseudo}. Besides improvements only on inference speed, there are other works aiming at speeding up both training and inference. Luhman et al.~\cite{luhman2022improving} propose a patch operation to decrease the dimensionality of each channel while accordingly increasing the number of channels, which greatly reduces the complexity of computation. Besides, a trainable forward process \cite{zhang2021diffusion} is also proven to benefit a faster training and inference speed. However, the price of their acceleration is a poor sampling quality evaluated by FID score. In this work, we accelerate DDPM on both training and inference from a different perspective by heavily reducing the dimensionality of the early diffusion process and thus improving the efficiency while obtaining on-par or even better quality of generation.


\vspace{2pt}
\noindent\textbf{Varying dimensionality of diffusion models.}
Due to the redundancy in image signals, it is possible to improve the efficiency of diffusion models by varying dimensionality during the generation process.
The most relevant work to our proposed model is subspace diffusion~\cite{jing2022subspace}, which can also vary dimensionality in the diffusion process.
However, subspace diffusion suffers from a trade-off between sampling acceleration and sample quality as claimed in~\cite{jing2022subspace}, while our \methodabbr can relieve this dilemma (see theoretical analysis in \cref{subsec:method-compare} and experimental results in \cref{subsec:exp-comparison}). 
Instead of varying dimensionality in one diffusion process, there are works cascading several diffusion processes with growing dimensionality \cite{ho2022cascaded, ramesh2022hierarchical, saharia2022photorealistic, ryu2022pyramidal}, where the subsequent process is conditioned on the previous samples.

\vspace{2pt}
\noindent\textbf{Discussion with latent diffusion.}
Besides varying dimensionality in image space, there are other methods, which we generally call latent diffusion, that directly apply diffusion models in a low dimensional latent space, obtained by an autoencoder~\cite{hu2022global, rombach2022high, esser2021imagebart, vahdat2021score, preechakul2022diffusion}. Although latent diffusion can also speed up the training and sampling of diffusion models, it decreases dimensionality by an additional model and keeps the diffusion process unchanged. In this paper, however, we focus on the improvement on the diffusion process itself to accelerate training and sampling, which is totally a different route. Besides training and sampling efficiency, another important contribution of this work is to prove that it it unnecessary for diffusion process to keep a fixed dimension along time. By controlling the attenuation of each data component, it is possible to change dimensionality while keeping the process reversible. Thus, we will not further compare our \methodabbr with latent diffusion.
\section{Background}\label{sec:background}

We first introduce the background of Denoising Diffusion Probabilistic Model (DDPM) \cite{sohl2015deep, ho2020denoising} and some of its extensions which are closely related to our work. DDPM constructs a forward process to perturb the distribution of data $q(\rvx_0)$ into a standard Gaussian $\gN(\mathbf{0}; \mI)$. Considering an increasing variance schedule of noises $\beta_1, \dots, \beta_T$, DDPMs define the forward process as a Markov chain
\begin{equation}
\vspace{-5pt}
    \rvx_t = \sqrt{1 - \beta_t} \rvx_{t - 1} + \sqrt{\beta_t} \rvepsilon,~t=1,2,\cdots,T,
\end{equation}
where $\rvepsilon$ is a standard Gaussian noise. In order to generate high-fidelity images, DDPM \cite{ho2020denoising} denoises samples from a standard Gaussian iteratively utilizing the reverse process parameterized as 
\begin{equation}
\vspace{-5pt}
    \rvx_{t-1} = \frac{1}{\sqrt{\alpha_t}}\left(\rvx_t-\frac{\beta_t}{\sqrt{1-\bar{\alpha}_t}}\epsilon_\theta(\rvx_t, t)\right) + \sqrt{\beta_t} \rvepsilon,
\end{equation}
where $\alpha_t = 1 - \beta_t,~\bar{\alpha}_t = \prod_{s=1}^t \alpha_s$, and $\rvepsilon_\theta$ is a neural network used to predict $\rvepsilon$ from $\rvx_t$. The parameters $\theta$ are learned by minimizing the following loss function
\begin{equation}
\vspace{-5pt}
    L(\theta) = \mathbb{E}_{t,\rvx_0,\rvepsilon} \left[\left\Vert 
 \rvepsilon - \rvepsilon_\theta(\rvx_t(\rvx_0, \rvepsilon), t) \right\Vert^2 \right].
\end{equation}
The standard diffusion model is implemented directly in the image space, which is probably not the optimal choice according to \cite{lee2022progressive}, and the relative importance of different frequency components can be taken into consideration. \cite{lee2022progressive} implements the diffusion models in a designed space by generalizing diffusion process with the forward process formulated as 
\begin{equation}
\vspace{-5pt}
    \rvx_t = \mU (\mI - \mB_t)^{\frac{1}{2}} \mU^T \rvx_{t-1} + \mU \mB_t \mU^T \rvepsilon,
\end{equation}
where $\mU$ is an orthogonal matrix to impose a rotation on $\rvx_t$, and the noise schedule is defined by the diagonal matrix $\mathbf{B}_t$. In this work, we extend the aforementioned generalized framework further and make it possible to vary dimensionality during the diffusion process.
\begin{figure*}[t]
\begin{center}
\includegraphics[width=.85\textwidth]{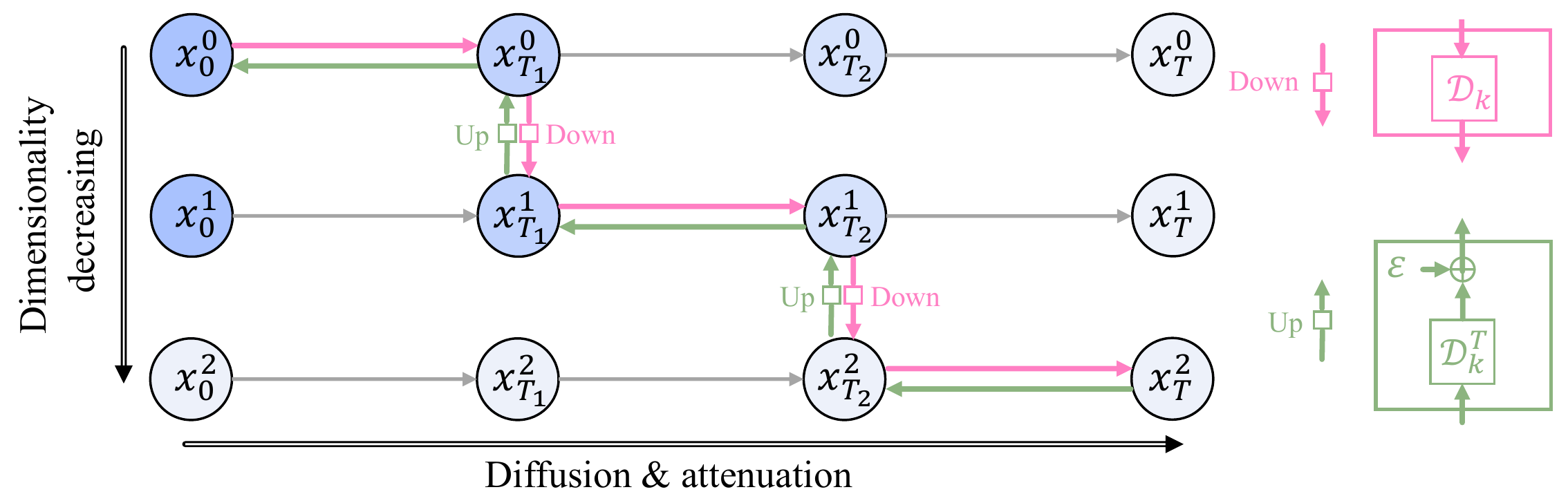}
\end{center}
\vspace{-15pt}
\caption{
    \textbf{Framework illustration} of \methodabbr.
    Each row represents an \Adp (\adpabbr), which controls the attenuation of each data component while adding noise. All $K+1$ {\adpabbr}s ($K=2$ here) have different dimensionality decreasing from top to bottom, and are concatenated by some simple opeartions to obtain our \methodabbr. In the forward process, the concatenation is achieved by \textit{downsampling operation}, and in reverse process, it is the \textit{upsampling operation} followed by adding a Gaussian noise.
}
\label{fig:our-diffusion}
\vspace{-5pt}
\end{figure*}

\section{\Method}\label{sec:method}

We formulate the \method(\methodabbr) in this section, which progressively decreases the dimension of $\rvx_t$ in forward process, and can be effectively reversed to generate high-dimensional data from a low-dimensional noise.
To establish \methodabbr, we gradually attenuate components of $\rvx_0$ in different subspaces and decrease the dimensionality of $\rvx_t$ at dimensionality turning points by downsampling operator (\cref{subsec:method-forward}), which is approximately reversible (\cref{subsec:method-reverse}) with controllable small error caused by the loss of attenuated $\rvx_0$ component (\cref{subsec:method-error}).

\subsection{Forward Process of \methodabbr}\label{subsec:method-forward}

In this section, we will construct the forward process of our \methodabbr, which decreases the dimensionality as time evolves and can be effectively reversed.
To this end, we concatenate multiple diffusion processes with different dimensions into an entire Markov chain by downsampling operations, while we elaborately design each process so that the information loss induced by downsampling is negligible. \cref{fig:our-diffusion} illustrates this overall framework. 
The concatenation of different processes enables us to decrease the dimensionality, and the control on information loss ensures that downsampling operations are approximately reversible, such that the entire process can be reversed (discussed later in \cref{subsec:method-reverse}).

To limit the information loss, we decompose the data into orthogonal components and control the attenuation of each component in the forward process of each concatenated diffusion, which we call \textit{Attenuated Diffusion Process} (\adpabbr).
Once the lost data component induced by downsampling is small enough, the information loss will be negligible.
In the following part of this section, we will first introduce the design of each \adpabbr.
Then we will show how to merge these {\adpabbr}s to obtain our \methodabbr.

\vspace{5pt}
\noindent\textbf{Notation list.}
We first define a sequence of subspaces and other necessary notations as follows:
\begin{itemize}
\vspace{-2pt}
\setlength{\itemsep}{0pt}
\setlength{\parsep}{0pt}

    \item $\sS_0 \supsetneq \sS_1 \supsetneq \cdots \supsetneq \sS_K$ is a sequence of subspaces with decreasing dimensionality $d=\bar{d}_0 > \bar{d}_1 > \cdots > \bar{d}_K$, where $\sS_0 = \sR^{d}$ is the original space, $K\in \sN_+$. For simplicity, $\sS_{K+1} \defeq \{\vzero\}$ and $\bar{d}_{K+1} \defeq 0$.
    
    \item $d_i = {\rm dim}(\sS_i / \sS_{i+1}),~i=0,1,\cdots,K$. Note that $d_K = {\rm dim}(\sS_K / \sS_{K+1}) = {\rm dim}(\sS_K)$.
    
    \item $\hat{\mU}_i \in \sR^{d \times d_i}$ is a matrix whose columns span subspace $\sS_{i}/\sS_{i+1}$ for $i=0,1,\cdots,K$.
    
    \item $\mU_0 = [\hat{\mU}_0,\cdots,\hat{\mU}_K] \in \sR^{d\times d}$ is an orthogonal matrix.
    
    \item $\mU_k \in \sR^{\bar{d}_k\times \bar{d}_k}$ is orthogonal matrix, $k=1,2,\cdots,K$.
    
    \item $\mU_k = [\mN_k, \mB_k]$ splits each $\mU_k$ into two sub-matrices, where $\mN_k\in \sR^{\bar{d}_k \times d_k}$, $\mB_k \in \sR^{\bar{d}_k \times \bar{d}_{k+1}}$, $k=0,1,\cdots,K$.
    
    \item $\mI_n \in \sR^{n\times n}$ is an identity matrix.
    
    \item $\mO_n \in \sR^{n\times n}$ is a zero matrix.
\end{itemize}

With the above definitions, we can first construct an \adpabbr $\rvx_t^0,~t=0,1,\cdots,T$ in $\sS_0$ as
\begin{equation}
\vspace{-5pt}
\begin{aligned}
\label{eq:xt0-given-x0}
    \rvx_t^0 =& \sum_{i=0}^K (\bar{\lambda}_{i,t} \rvv_i^0 + \bar{\sigma}_{i,t} \rvz_i^0)\\
             =& \mU_0 \bar{\mLambda}_{0,t} \mU_0^\T \rvx_0^0 + \mU_0 \bar{\mL}_{0,t} \mU_0^\T \rvepsilon^0, 
\end{aligned}
\end{equation}
where $\rvv_i^0 \in \sS_i / \sS_{i+1}$ is the component of original data point $\rvx_0^0$ in subspace $\sS_i / \sS_{i+1}$,
$\bar{\lambda}_{i,t}$ controls the attenuation of $\rvv_i^0$ along timestep $t$,
$\rvz_i^0$ is the component of a standard Gaussian noise $\rvepsilon^0 \in \sR^{\bar{d}_0}$ in the same subspace as $\rvv_i^0$ (\textit{i.e.}, $\sS_i / \sS_{i+1}$),
$\bar{\sigma}_{i,t}$ is the standard deviation of $\rvz_i^0$,
and $\bar{\mLambda}_{0,t},~\bar{\mL}_{0,t} \in \sR^{\bar{d}_0 \times \bar{d}_0}$ are two diagonal matrices defined as $\bar{\mLambda}_{0,t}={\rm diag}(\bar{\lambda}_{0,t}\mI_{d_0},~\bar{\lambda}_{1,t}\mI_{d_1},~\cdots,~\bar{\lambda}_{K,t}\mI_{d_K})$, $\bar{\mL}_{0,t}={\rm diag}(\bar{\sigma}_{0,t}\mI_{d_0},~\bar{\sigma}_{1,t}\mI_{d_1},~\cdots,~\bar{\sigma}_{K,t}\mI_{d_K})$.
For compatibility at $t=0$, we have $\bar{\mLambda}_{0,0} = \mI_{\bar{d}_0}$ and $\bar{\mL}_{0,0} = \mO_{\bar{d}_0}$, \textit{i.e.}, $\bar{\lambda}_{i,0}=1$ and $\bar{\sigma}_{i,0}=0$ for all $i=0,1,\cdots,K$.
To control the attenuation of each data component $\rvv_i^0$, we require $\bar{\lambda}_{i,t}$ to gradually decrease from 1 to approximate 0 for $i=0,1,\cdots,K-1$ as timestep $t$ evolves. As for $\bar{\lambda}_{K,t}$, it is not required to decrease (explained later after \cref{eq:dvdp-chain}).

Starting from $\rvx_t^0$, we can recursively construct a dimensionality-decreasing sequence of {\adpabbr}s
\begin{equation}
\vspace{-5pt}
\begin{aligned}
\label{eq:xtk-given-x0}
    \rvx_t^k &= \gD_k \rvx_t^{k - 1} = \sum_{i=k}^K (\bar{\lambda}_{i,t} \rvv_i^k + \bar{\sigma}_{i,t} \rvz_i^k)\\
             &= \mU_k \bar{\mLambda}_{k,t} \mU_k^\T \rvx_0^k + \mU_k \bar{\mL}_{k,t} \mU_k^\T \rvepsilon^k,~1\leq k \leq K,
\end{aligned}
\end{equation}
where $\gD_k: \sR^{\bar{d}_{k-1}} \rightarrow \sR^{\bar{d}_k}$ is a linear surjection, which we call the $k$-th \textit{downsampling operator} as it reduces the dimensionality of the operand (without ambiguity, we also use $\gD_k$ to denote the corresponding matrix in $\sR^{\bar{d}_k \times \bar{d}_{k-1}}$),
$\rvv_i^k = \gD_k \rvv_i^{k-1} \in \overline{\gD}_k (\sS_i / \sS_{i+1})$ is the component of $\rvx_0^k = \gD_k \rvx_o^{k-1} \in \sR^{\bar{d}_k}$ ($\overline{\gD}_k \defeq \prod_{i=1}^k \gD_i$),
 $\rvz_i^k = \gD_k \rvz_i^{k-1} \in \overline{\gD}_k (\sS_i / \sS_{i+1})$ is the component of a standard Gaussian noise $\rvepsilon^k = \gD_k \rvepsilon^{k-1} \in \sR^{\bar{d}_k}$,
$\bar{\mLambda}_{k,t}={\rm diag}(\bar{\lambda}_{k,t}\mI_{d_k},~\bar{\lambda}_{k+1,t}\mI_{d_{k+1}},~\cdots,~\bar{\lambda}_{K,t}\mI_{d_K}) \in \sR^{\bar{d}_k \times \bar{d}_k}$,
$\bar{\mL}_{k,t}={\rm diag}(\bar{\sigma}_{k,t}\mI_{d_k},~\bar{\sigma}_{k+1,t}\mI_{d_{k+1}},~\cdots,~\bar{\sigma}_{K,t}\mI_{d_K}) \in \sR^{\bar{d}_k \times \bar{d}_k}$,
and orthogonal matrix $\mU_k \in \sR^{\bar{d}_k \times \bar{d}_k}$ satisfies $\mU_k = \gD_k \mB_{k-1}$ (see \textit{notation list} for the definition of $\mB_{k-1}$).
From \cref{eq:xtk-given-xt-1}, it is clear that components $\rvv_{k-1}^{k-1}$ and $\rvz_{k-1}^{k-1}$ will be lost every time $\gD_k$ is applied on $\rvx_{t}^{k-1}$, which further requires $\gD_k$ to satisfy $\gD_k \mN_{k-1} = \bm{0}$ (see \textit{notation list} for the definition of $\mN_{k-1}$).
Both \cref{eq:xt0-given-x0,eq:xtk-given-x0} can be derived from Markov chains with Gaussian kernels as (see \Cref{subsec:append-forward-transition} for the proof)
\begin{equation}
\label{eq:xtk-given-xt-1}
    \rvx_t^k = \mU_k \mLambda_{k,t} \mU_k^\T \rvx_{t-1}^k + \mU_k \mL_{k,t} \mU_k^\T \rvepsilon^k,~0\leq k \leq K,
\end{equation}
where $\mLambda_{k,t}\!=\! \bar{\mLambda}_{k,t-1}^{-1} \bar{\mLambda}_{k,t}$, $\mL_{k,t} \!=\! (\bar{\mL}_{k,t}^2 - \mLambda_{k,t}^2 \bar{\mL}_{k,t-1}^2 )^{1/2}$.

Now with the {\adpabbr}s $\{\rvx_t^k\}_{k=0}^K$ given by \cref{eq:xtk-given-xt-1}, we can construct the forward process of our \methodabbr by merging different parts of $\{\rvx_t^k\}_{k=0}^K$ in the following manner:
consider a strictly increasing time sequence $T_1, T_2, \cdots, T_K$, if for each $k$ satisfying $1 \leq k \leq K$, $\bar{\lambda}_{k-1,T_k}$ becomes small enough, then $\rvx_{T_k}^{k-1}$ is downsampled by $\gD_k$ to obtain $\rvx_{T_k}^k$ with lower dimensionality, and each $T_k$ is a \textit{dimensionality turning point}.
The entire process can be expressed as
\begin{equation}
\vspace{-5pt}
\begin{alignedat}{5}
\label{eq:dvdp-chain}
&\rvx_0^0 &&\longrightarrow \rvx_1^0 &&\longrightarrow &&\cdots &&\longrightarrow \rvx_{T_1}^0\\
         \overset{\gD_1}{\longrightarrow} &\rvx_{T_1}^1 &&\longrightarrow \rvx_{T_1+1}^1 &&\longrightarrow &&\cdots &&\longrightarrow \rvx_{T_2}^1\\
         \vdots \quad &&& \\
         \overset{\gD_K}{\longrightarrow} &\rvx_{T_K}^K &&\longrightarrow \rvx_{T_K + 1}^K &&\longrightarrow &&\cdots &&\longrightarrow \rvx_{T}^K
\end{alignedat}
\end{equation}
We further explain this process as follows:
\begin{itemize}
    \setlength{\itemsep}{0pt}
    \item Between two adjacent dimensionality turning points $T_{k-1}$ and $T_{k}$, $\rvx_t^{k-1}$ diffuses and attenuates data components $\rvv_i^{k-1},~i \geq k-1$, which keeps the dimensionality $\bar{d}_{k-1}$.
    The attenuation of $\rvv_i^{k-1}$ is achieved by the time-decreasing coefficient $\bar{\lambda}_{i, t}$.
    \item When it comes to $T_k$, $\rvx_{T_k}^{k-1} \overset{\gD_k}{\longrightarrow} \rvx_{T_k}^k$ decreases the dimension from $\bar{d}_{k-1}$ to $\bar{d}_k$.
    \item After the last dimensionality turning point $T_k$, $\rvx_t^K$ can just evolve as conventional diffusion without data component attenuation by keeping a constant $\bar{\lambda}_{K, t}$.
\end{itemize}
Thus, the entire process in \cref{eq:dvdp-chain} decreases the dimensionality by $K$ times from $\bar{d}_0 = d$ to $\bar{d}_K$.
It should be noted that process in \cref{eq:dvdp-chain} is also Markovian since each diffusion sub-process $\rvx_{T_{k-1}}^{k-1} \rightarrow \rvx_{T_k}^{k-1}$ is Markovian, and the result of each downsampling operation $\rvx_{T_k}^k$ is uniquely determined by the previous state $\rvx_{T_k}^{k-1}$.
Also note that each downsampling operation $\gD_k$ loses little information because of small $\bar{\lambda}_{k-1,T_k}$ which is a controllable hyperparameter.
For better understanding, consider the relationship between $\rvx_{T_k}^{k-1}$ and $\rvx_{T_k}^k$ derived from \cref{eq:xtk-given-x0}
\begin{equation}
\vspace{-5pt}
\label{eq:downsample-loss}
\rvx_{T_k}^{k-1} = \gD_k^T \rvx_{T_k}^k + \underbrace{\bar{\lambda}_{k-1, T_k} \rvv_{k-1}^{k-1}}_{\text{data component}} + \underbrace{\bar{\sigma}_{k-1, T_k} \rvz_{k-1}^{k-1}}_{\text{noise component}}.
\end{equation}
where $\gD_k^\T$ is the transpose of matrix $\gD_k$, named as the $k$-th \textit{upsampling operator}.
From \cref{eq:downsample-loss}, it is clear that $\rvx_{T_k}^k$ actually loses two terms compared with $\rvx_{T_k}^{k-1}$: 1) data component $\bar{\lambda}_{k-1, T_k} \rvv_{k-1}^{k-1}$ which is informative but negligible as $\bar{\lambda}_{k-1, T_k}$ is set to be small enough, and 2) noise component $\bar{\sigma}_{k-1, T_k} \rvz_{k-1}^{k-1}$ that is non-informative and can be compensated in the reverse process, as we will discuss in \cref{subsec:method-reverse}.

\subsection{Reverse Process Approximating \methodabbr}\label{subsec:method-reverse}

In this section, we will derive an approximate reverse process, which induces a data generation process with progressively growing dimensionality.
The approximation error will be discussed in \cref{subsec:method-error}, and we can find that it actually converges to zero. 
Loss function will also be given at the end of this section.
Implementation details of training and sampling can be found in \Cref{sec:append-implement}.

\vspace{5pt}
\noindent\textbf{Reverse transition.}
Since \methodabbr is a sequence of fixed-dimensionality diffsion processes connected by downsampling operations at dimensionality turning points, we consider reverse transition kernels \textit{between} and \textit{at} dimensionality turning points separately.

For reverse transition \textit{between} two adjacent dimensionality turning points, \textit{i.e.}, $p_\theta(\rvx_{t-1}^k | \rvx_{t}^k)$ with $T_{k-1} \leq t \leq T_k$ for $k \in [1, K]$, it can be defined as a Gaussian kernel $p_\theta(\rvx_{t-1}^k | \rvx_{t}^k) = \gN(\rvx_{t-1}^k; \vmu_\theta(\rvx_t^k, t), \mSigma_t)$.
As in DDPM~\cite{ho2020denoising}, the reverse process covariance matrices $\mSigma_t$ are set to untrained time-dependent constants, and the mean term $\vmu_\theta(\rvx_t^k, t)$ is defined as (see \Cref{subsec:append-forward-post} for details)
\begin{equation}
\vspace{-5pt}
\begin{aligned}
\label{eq:reverse-mean}
    \vmu_\theta =& \tilde{\vmu}_{k,t} \big( \rvx_t^k, \mU_k \bar{\mLambda}_{k,t}^{-1} \mU_k^\T \rvx_t^k \\
    &- \mU_k \bar{\mLambda}_{k,t}^{-1} \bar{\mL}_{k,t} \mU_k^\T \veps_\theta(\rvx_t^k, t) \big),
\end{aligned}
\end{equation}
where $\tilde{\vmu}_{k,t}$ is the mean function of forward process posterior $q(\rvx_{t-1}^k | \rvx_t^k, \rvx_0^k) = \gN(\rvx_{t-1}^k; \tilde{\vmu}_{k,t}(\rvx_t^k, \rvx_0^k), \tilde{\mSigma}_{k,t})$, and $\veps_\theta$ represents a trainable network.

For reverse transition \textit{at} dimensionality turning points, \textit{i.e.}, $p_\theta(\rvx_{T_k}^{k-1} | \rvx_{T_k}^k)$ for $k \in [1, K]$, the corresponding forward transitions barely lose information as illustrated by \cref{eq:downsample-loss} in \cref{subsec:method-forward}, thus $\rvx_{T_k}^k \rightarrow \rvx_{T_k}^{k-1}$ can be approximately achieved without any trainable network as
\begin{equation}
\label{eq:reverse-up}
    \rvx_{T_k}^{k-1} = \gD_k^\T \rvx_{T_k}^k + \mU_{k-1} \Delta \mL_{k-1} \mU_{k-1}^\T \rvepsilon^{k-1},
\end{equation}
where $\gD_k^\T \in \sR^{\bar{d}_k \times \bar{d}_{k-1}}$ is the upsampling operator, and $\Delta \mL_{k-1} = {\rm diag}(\bar{\sigma}_{k-1, T_k}\mI_{d_{k-1}}, \mO_{\bar{d}_k})$ represents the standard deviation of added Gaussian noise. \cref{eq:reverse-up} can be understood as: we first upsample $\rvx_{T_k}^{k}$, then compensate for a Gaussian noise with the same covariance as the lost noise component in the forward downsampling operation, \textit{i.e.}, $\bar{\sigma}_{k-1, T_k} \rvz_{k-1}^{k-1}$ in \cref{eq:downsample-loss}. The approximation error comes from neglecting data component $\bar{\lambda}_{k-1, T_k}\rvv_{k-1}^{k-1}$, and will be analyzed later in \cref{subsec:method-error}.

\vspace{5pt}
\noindent\textbf{Loss function.} Similar with DDPM~\cite{ho2020denoising}, a loss function can be derived from a weighted variational bound as (see \Cref{subsec:append-loss} for details)
\begin{equation}
\hspace{-3pt}
\label{eq:reverse-loss}
    L(\theta) = \mathbb{E}_{k} \mathbb{E}_{\rvx_0^k, \rvepsilon^k, t\sim \gU_k}\big[\big\Vert 
 \rvepsilon^k - \rvepsilon_\theta(\rvx_{t}^k(\rvx_0^k, \rvepsilon^k), t) \big\Vert^2 \big],
\end{equation}
where $\gU_k = \gU\big((T_k, T_{k+1}]\big)$ is a discrete uniform distribution between $T_k$ (exclusive) and $T_{k+1}$ (inclusive), and $\rvx_{t}^k(\rvx_0^k, \rvepsilon^k)$ represents the forward $\rvx_{t}^k$ determined by $\rvx_0^k$ and $\rvepsilon^k$ given in \cref{eq:xtk-given-x0}.

\subsection{Error Analysis}\label{subsec:method-error}

In \cref{subsec:method-reverse}, we mention that the reverse process is just an approximation of the forward \methodabbr at each dimensionality turning point $T_k$.
In this section, we will measure this approximation error in probability sense, \textit{i.e.}, the difference between the real forward distribution $q(\rvx_{T_k}^{k-1})$ and the reverse distribution $p(\rvx_{T_k}^{k-1})$ under proper assumptions, and will find that this difference converges to zero.

To measure the difference between two distributions, we use \textit{Jensen-Shannon Divergence} (JSD) as a metric.
Under this metric, upper bound of the approximation error can be derived from \Cref{prop:xt-diff} (see \Cref{subsec:append-proof-prop1} for proof):
\begin{proposition}
\vspace{-3pt}
\label{prop:xt-diff}
Assume $p_1(\vx | \vx_0)$, $p_2(\vx | \vx_0)$ are two Gaussians such that $p_1(\vx | \vx_0) = \mathcal{N}(\vx ; \mA_1 \vx_0, \mSigma )$ and $p_2(\vx | \vx_0) = \mathcal{N}(\vx ;\mA_2 \vx_0, \mSigma )$, where positive semi-definite matrices $\mA_1,~\mA_2$ satisfies $\mA_1 \succeq \mA_2 \succeq 0$, covariance matrix $\mSigma$ is positive definite, and the support of distribution $p(\vx_0)$ is bounded, then Jensen-Shannon Divergence ({\rm JSD}) of the two marginal distributions $p_1(\vx)$ and $p_2(\vx)$ satisfies
\begin{equation}
\label{eq:prop-error-bound}
\begin{aligned}
    {\rm JSD}(p_1 || p_2) \leq& \frac{\sqrt{2}}{2} e^{-\frac{1}{2}} B \left(2\sqrt{2} + \frac{V_d(r)}{(2\pi)^{\frac{d}{2}}}\right) \\
    &\cdot \Vert \mSigma^{-\frac{1}{2}} (\mA_1 - \mA_2)\Vert_2,
\end{aligned}
\end{equation}
where $B$ is the upper bound of $\Vert \vx_0\Vert_2$, $V_d(\cdot)$ is the volume of $d$-dimensional sphere with respect to the radius, and $r = 2 B \Vert \mSigma^{-\frac{1}{2}} A_1\Vert_2$.
\vspace{-5pt}
\end{proposition}

With \Cref{prop:xt-diff}, the upper bound of JSD between the forward distribution $q(\rvx_{T_k}^{k-1})$ and the reverse distribution $p(\rvx_{T_k}^{k-1})$ can be obtained by \Cref{thm:reverse-error} (see \Cref{subsec:append-proof-thm1} for proof)
\begin{theorem}[Reverse Process Error]
\vspace{-5pt}
\label{thm:reverse-error}
Assume $0 < k \leq K$, $q(\rvx_{T_k}^{k-1})$ and  $q(\rvx_{T_k}^k)$ are defined by \cref{eq:xt0-given-x0,eq:xtk-given-x0}, $p(\rvx_{T_k}^{k-1})$ is the marginal distribution of $q(\rvx_{T_k}^k)p(\rvx_{T_k}^{k-1} | \rvx_{T_k}^k)$ where $p(\rvx_{T_k}^{k-1} | \rvx_{T_k}^k)$ is defined by \cref{eq:reverse-up}, and $\Vert \rvx_0 \Vert_2 \leq \sqrt{d}$, then
\begin{equation}
\label{eq:thm1-error-bound}
\begin{aligned}
    \xi_1 \leq& \frac{\sqrt{2}}{2} e^{-\frac{1}{2}} \sqrt{d} \left(2\sqrt{2} + \frac{V_d(r)}{(2\pi)^{\frac{d}{2}}}\right) \frac{\bar{\lambda}_{k-1,T_k}}{\bar{\sigma}_{k-1,T_k}}\\
    =& o(\bar{\lambda}_{k-1,T_k})
\end{aligned}
\end{equation}
where $\xi_1\defeq {\rm JSD}(q(\rvx_{T_k}^{k-1}) || p(\rvx_{T_k}^{k-1}))$, and $r = 2 \sqrt{d} \max_{k-1 \leq i \leq K}\frac{\bar{\lambda}_{i,T_k}}{\bar{\sigma}_{i,T_k}}$.
\vspace{-5pt}
\end{theorem}
Note that the assumption $\Vert \rvx_0 \Vert_2 \leq \sqrt{d}$ can be satisfied for image data, since pixel values can be normalized in $[-1, 1]$. Thus, \Cref{thm:reverse-error} claims that $\xi_1$ can be arbitrarily small as $\bar{\lambda}_{k-1, T_k} \rightarrow 0$ if we can get an exact $q(\rvx_{T_k}^{k-1})$ by reverse process. It means that the approximation error caused by stepping over $T_k$ can be small enough.

\subsection{Comparison with Subspace Diffusion}\label{subsec:method-compare} 

To reduce the dimensionality of latent space in diffusion models, subspace diffusion is proposed to model in a low-dimensional subspace at high noise levels, and keep the original full-dimensional network at low noise levels~\cite{jing2022subspace}.
This can also be seen as a concatenation of multiple diffusion processes with different dimensionality like our \methodabbr, but without controllable attenuation on each data component. Each concatenated processes is just conventional isotropic diffusion.

Thus, subspace diffusion can be seen as a special case of our \methodabbr with $\bar{\lambda}_t \defeq \bar{\lambda}_{0,t} = \bar{\lambda}_{1,t} = \cdots = \bar{\lambda}_{K,t}$ and $\bar{\sigma}_t \defeq \bar{\sigma}_{0,t} = \bar{\sigma}_{1,t} = \cdots = \bar{\sigma}_{K,t}$, which limits the choice of dimensionality turning points.
This limitation can be further explained by \cref{eq:downsample-loss}: in the forward process, $\rvx_{T_k}^{k-1}$ will lose an informative data component $\bar{\lambda}_{k-1, T_k} \rvv_{k-1}^{k-1}$ and a non-informative noise component $\bar{\sigma}_{k-1, T_k} \rvz_{k-1}^{k-1}$ at dimensionality turning point $T_k$. To safely neglect the data component in the reverse transition, it requires that $\bar{\sigma}_{k-1, T_k} / \bar{\lambda}_{k-1, T_k} \gg \Vert \rvv_{k-1}^{k-1} \Vert / \Vert  \rvz_{k-1}^{k-1} \Vert$. 
For subspace diffusion, it means that the consistent $\bar{\sigma}_t / \bar{\lambda}_t$ for components in all subspaces should be high enough, which usually indicates a large $T_k$, \textit{i.e.}, a large number of diffusion steps in high dimensional space.

Therefore, as claimed in \cite{jing2022subspace}, the choice of $T_k$ should balance two factors: 1) smaller $T_k$ reduces the number of reverse diffusion steps occurring at higher dimensionality, whereas 2) larger $T_k$ makes the reverse transition at $T_k$ more accurate.
Although \cite{jing2022subspace} additionally proposes to compensate the loss of data component by adding an extra Gaussian noise besides compensation for the noise component, this trade-off still exists.
However, our \methodabbr can set much smaller $T_k$ with little loss in accuracy, which benefits from the controllable attenuation for each data component. \Cref{thm:reverse-error} supports this advantage theoretically, and experimental results in \cref{subsec:exp-comparison} further demonstrate it.

\begin{table*}[t]
\caption{
    \textbf{Quantitative comparison} between DDPM~\cite{ho2020denoising} and our \methodabbr on various datasets regarding image quality and model efficiency. $\ast$ indicates our reproduced DDPM. Both DDMP$^\ast$ and our \methodabbr adopt the improved UNet~\cite{dhariwal2021diffusion} for a fair comparison.
}
\label{tab:small-image}
\vspace{-15pt}
\begin{center}
\setlength{\tabcolsep}{4pt}
\begin{tabular}{llccccc}
\toprule
\multirow{2}{*}{Dataset}  &\multirow{2}{*}{Method} &\multirow{2}{*}{FID (50k)$\downarrow$} &Training Speed &Training Speed &Sampling Speed &Sampling Speed\\ & & & (sec/iter) & Up & (sec/sample) & Up\\
\midrule 
\multirow{3}{*}{CIFAR10 $32\times32$}
&DDPM         &$3.17$       &$-$            &$-$            &$-$            &$-$\\
&DDPM$^\ast$  &$\bf 3.16$   &$0.18$         &$-$            &$0.34$         &$-$\\
&\methodabbr  &$3.24$       &$\bf 0.15$     &$1.2\times$    &$\bf 0.26$     &$1.3\times$\\
\midrule 
\multirow{3}{*}{LSUN Bedroom $256\times256$}
&DDPM         &$6.36$       &$-$            &$-$            &$-$            &$-$\\
&DDPM$^\ast$  &$5.74$       &$0.99$         &$-$            &$12.2$         &$-$\\
&\methodabbr  &$\bf 4.88$   &$\bf 0.45$     &$2.2\times$    &$\bf 5.01$     &$2.4\times$\\
\midrule 
\multirow{3}{*}{LSUN Church $256\times256$}
&DDPM         &$7.89$       &$-$            &$-$            &$-$            &$-$\\
&DDPM$^\ast$  &$7.54$       &$0.99$         &$-$            &$12.2$         &$-$\\
&\methodabbr  &$\bf 7.03$   &$\bf 0.45$     &$2.2\times$    &$\bf 5.01$     &$2.4\times$\\
\midrule 
\multirow{3}{*}{LSUN Cat $256\times256$}
&DDPM         &$19.75$      &$-$            &$-$            &$-$            &$-$\\
&DDPM$^\ast$  &$18.11$      &$0.99$         &$-$            &$12.2$         &$-$\\
&\methodabbr  &$\bf 16.50$  &$\bf 0.45$     &$2.2\times$    &$\bf 5.01$     &$2.4\times$\\
\midrule 
\multirow{2}{*}{FFHQ $256\times256$}    
&DDPM$^\ast$  &$8.33$       &$0.99$         &$-$            &$12.2$         &$-$\\
&\methodabbr  &$\bf 8.03$   &$\bf 0.45$     &$2.2\times$    &$\bf 5.01$     &$2.4\times$\\
\bottomrule
\end{tabular}
\end{center}
\vspace{-10pt}
\end{table*}

\section{Experiments}\label{sec:exp}

In this section, we show that our \methodabbr can speed up both training and inference of diffusion models while achieving competitive performance. Besides, thanks to the varied dimension, \methodabbr is able to generate high-quality and high-resolution images from a low-dimensional subspace and exceeds existing methods including score-SDE~\cite{song2020score} and Cascaded Diffusion Models (CDM)~\cite{ho2022cascaded} on FFHQ $1024\times1024$. Specifically, we first introduce our experimental setup in \cref{subsec:exp-setup}. Then we compare our \methodabbr with existing alternatives on several widely evaluated datasets in terms of visual quality and modeling efficiency in \cref{subsec:exp-main-results}. After that, we compare our \methodabbr with Subspace Diffusion~\cite{jing2022subspace}, a closely related work proposed recently, in \cref{subsec:exp-comparison}. Finally, we implement the necessary ablation studies in the last \cref{subsec:exp-ablation}.

\subsection{Experimental Setup}\label{subsec:exp-setup}

\noindent\textbf{Datasets.}
In order to verify that \methodabbr is widely applicable, we use six image datasets covering various classes and a wide range of resolutions from 32 to 1024.
To be specific, we implement \methodabbr on CIFAR10 $32^2$~\cite{krizhevsky2009learning}, LSUN Bedroom $256^2$~\cite{yu2015lsun}, LSUN Church $256^2$, LSUN Cat $256^2$, FFHQ $256^2$ and FFHQ $1024^2$\cite{karras2019style}.

\vspace{2pt}
\noindent\textbf{Implementation details.}
%
We adopt the UNet improved by \cite{dhariwal2021diffusion} which achieves better performance than the traditional version \cite{ho2020denoising}.
Since most baseline methods adopt a single UNet network for all timesteps in the whole diffusion process, our \methodabbr also keeps this setting, except the comparison with subspace diffusion in \cref{subsec:exp-comparison}, which takes two networks for different generation stages~\cite{jing2022subspace}.
In principle, the network structure of our \methodabbr is kept the same as corresponding baseline.
However, when image resolution comes to $1024\times 1024$, the UNet in conventional diffusion models should be deep and contain sufficient downsampling blocks, thus to obtain embeddings with proper size (usually $4\times 4$ or $8\times 8$) in the bottleneck layer, while our \methodabbr does not need such a deep network since the generation starts from a low resolution noise ($64\times 64$ in our case).
Thus, for our \methodabbr, we only maintain a similar amount of parameters but use a different network structure from the baseline models.
We set the number of timesteps $T=1000$ in all of our experiments.
For \methodabbr, we reduce the dimensionality by $\frac{1}{4}$ (\textit{i.e.}, $h\times w \rightarrow \frac{h}{2}\times \frac{w}{2}$ for image resolution) when the timestep $t$ reaches one of the pre-designed dimensionality turning points, which are denoted as a set $\sT$.
For CIFAR10 $32 \times 32$, we set $\sT=\{600\}$, indicating that the resolution is decreased from $32\times 32$ to $16\times16$ when $t=600$.
Similarly, we set $\sT=\{300, 600\}$ for all $256\times256$ datasets and $\sT=\{200, 400, 600, 800\}$ for FFHQ $1024\times1024$.
In all of our experiments, the noise schedule of \methodabbr is an adapted version of linear schedule~\cite{ho2020denoising}, which is suitable for \methodabbr and keeps a comparable signal-to-noise ratio (SNR) with the original version (see \Cref{subsec:append-noise-schedule} for details).

\vspace{2pt}
\noindent\textbf{Evaluation metrics.}
For all of our experiments, we calculate the FID score \cite{heusel2017gans} of 50k samples to evaluate the visual quality of samples, except for FFHQ $1024\times1024$ with 10k samples due to a much slower sampling.
As for training and sampling speed, both of them are evaluated on a single NVIDIA A100 GPU. Training speed is measured by the mean time of each iteration (estimated over 4,000 iterations), and sampling speed is measured by the mean time of each sample (estimated over 100 batches).
The training batch size and sampling batch size are 128, 256 respectively for CIFAR10, and 24, 64 respectively for other 256$\times$256 datasets.

\subsection{Improving Visual Quality and Modeling Efficiency.}\label{subsec:exp-main-results}

\noindent\textbf{Comparison with existing alternatives.}
We compare \methodabbr with other alternatives here to show that \methodabbr has the capability of acceleration while maintaining a reasonable or even better visual quality. For the sake of fairness, we reproduce DDPM using the same network structure as \methodabbr with the same hyperparameters, represented as DDPM$^*$. \cref{tab:small-image} demonstrates our experimental results on CIFAR10, FFHQ $256\times256$, and three LSUN datasets. The results show that our proposed \methodabbr achieves better FID scores on all of the $256\times256$ datasets, illustrating an improved visual quality. Meanwhile, \methodabbr enjoys improved training and sampling speeds. Specifically, DDPM and DDPM$^*$ spend $2.2\times$ time as \methodabbr when training the same epochs, and they spend $2.4\times$ time generating one image. Although the superiority of \methodabbr is obvious on the $256\times256$ datasets, it becomes indistinct when it comes to CIFAR10 $32\times 32$, which is reasonable considering the negligible redundancy of images in CIFAR10 due to the low resolution.

\vspace{2pt}
\noindent\textbf{Towards high-resolution image synthesis.}
Because of the high computation cost, it is hard for diffusion models to generate high-resolution images. Score-SDE \cite{song2020score} tries this task by directly training a single diffusion model but the sample quality is far from reasonable. Recently, CDM attracts great interests in high-resolution image synthesis and obtains impressive results \cite{ramesh2022hierarchical, saharia2022photorealistic}. It generates low-resolution images by the first diffusion model, followed by several conditional diffusion models as super-resolution modules. We compare \methodabbr with score-SDE and CDM on FFHQ $1024\times1024$ in \cref{tab:fid-ffhq1024}. The FID of score-SDE is evaluated on samples generated from their official code and model weight without acceleration, and CDM is implemented by three cascaded diffusions as in \cite{ramesh2022hierarchical}. Our \methodabbr is sampled by both DDPM method with 1000 steps and DDIM method with 675 steps. The results show that \methodabbr beats both score-SDE and CDM.

\subsection{Comparison with Subspace Diffusion}\label{subsec:exp-comparison}

Subspace diffusion \cite{jing2022subspace} can also vary dimensionality during the diffusion process.
As mentioned in~\cite{jing2022subspace} and also discussed in \cref{subsec:method-compare}, dimensionality turning point $T_k$ in subspace diffusion should be large enough to maintain the sample quality.
However, large $T_k$ means more diffusion steps in high dimensional space, which will impair the advantage of such dimensisonality-varying method, \textit{e.g.}, less acceleration in sampling.
Thus, $T_k$ is expected to be as small as possible while maintaining the sampling quality.

\begin{figure*}[t]
\centering
\begin{minipage}[t]{0.45\textwidth}
\centering
\includegraphics[width=0.95\textwidth]{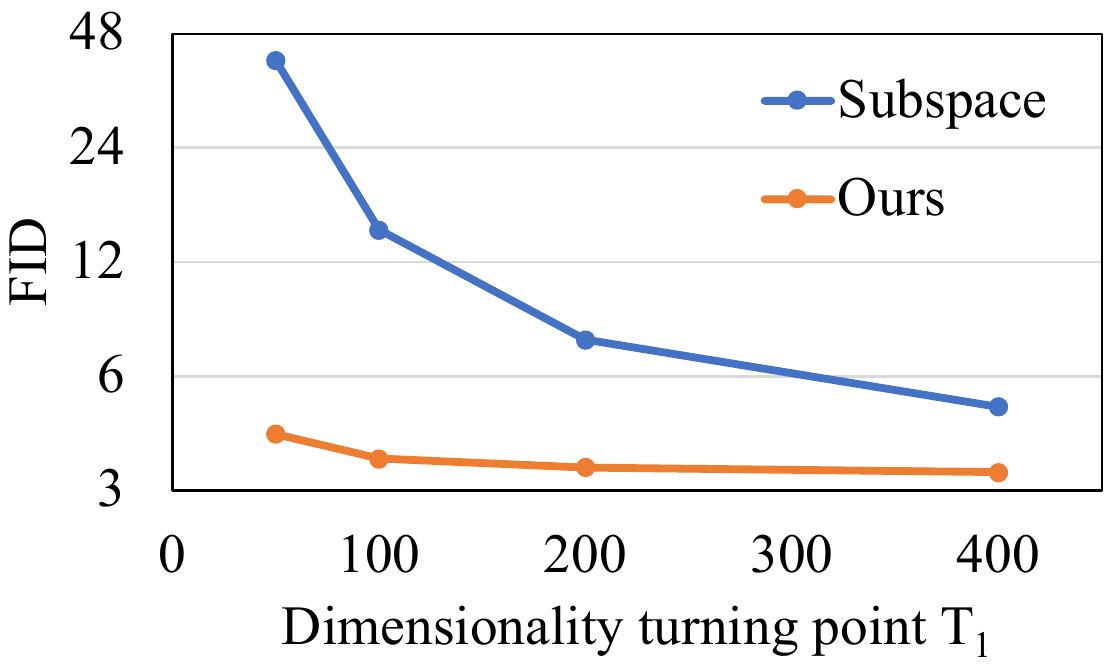}
\vspace{-5pt}
\caption{
    \textbf{Quantitative comparison} between subspace diffusion~\cite{jing2022subspace} and our \methodabbr on CelebA 64$\times$64 regarding different dimensionality turning point $T_1$.
}
\label{fig:subspace-fid-comparison}
\end{minipage}
\hfill
\begin{minipage}[t]{0.52\textwidth}
\centering
\includegraphics[width=0.95\textwidth]{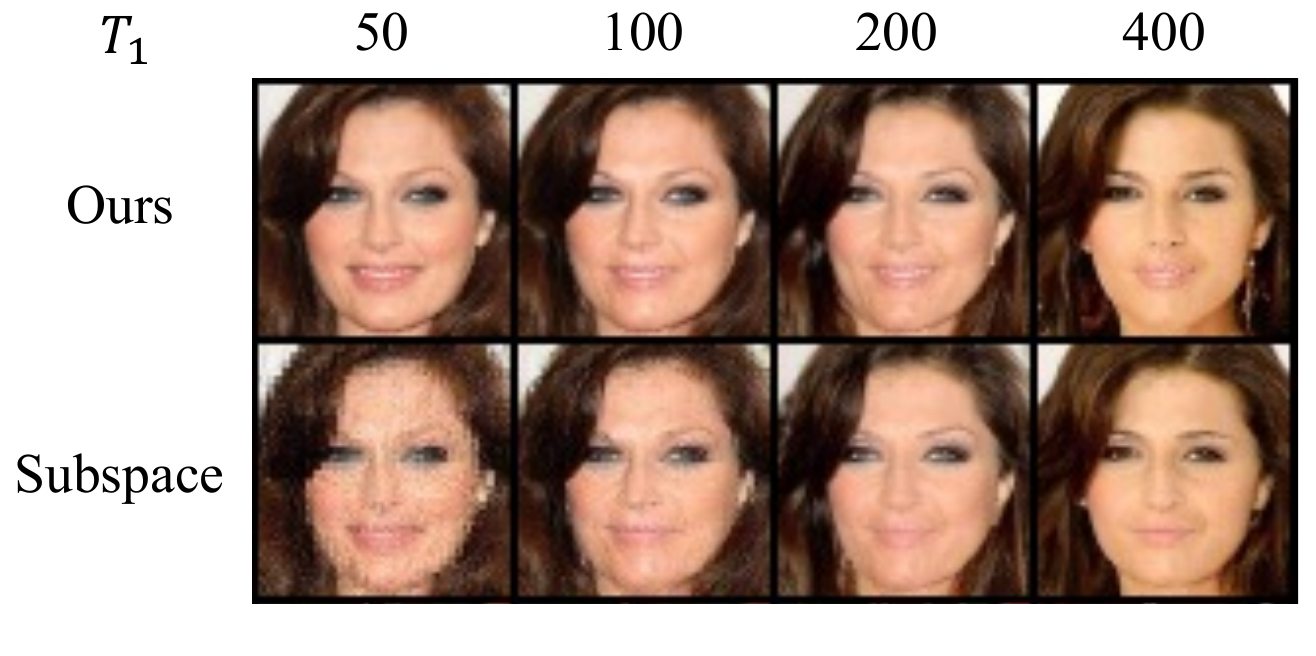}
\vspace{-5pt}
\caption{
    \textbf{Qualitative comparison} between subspace diffusion~\cite{jing2022subspace} and our \methodabbr on CelebA 64$\times$64.
    $T_1$ denotes the dimensionality turning point.
}
\label{fig:subspace-sample-comparison}
\end{minipage}
\vspace{-5pt}
\end{figure*}

Considering that the dimensionality decreases only once, \textit{i.e.}, $K=1$, and only one dimensionality turning point $T_1$, we compare \methodabbr with subspace diffusion when the downsampling is carried out at different $T_1$. Besides, since the subspace diffusion is only implemented on continuous timesteps before, we reproduce it on discrete timesteps similar as DDPM and use the reproduced version as a baseline. \cref{fig:subspace-fid-comparison} illustrates that \methodabbr is consistently better with regard to sample quality on CelebA $64\times64$ \cite{liu2015deep} when $T_1$ varies, where the advantage gets larger when $T_1$ gets smaller. In addition, some samples of \methodabbr and subspace diffusion are shown in \cref{fig:subspace-sample-comparison}, where the sample quality of subspace diffusion is apparently worse than that of \methodabbr especially when $T_1$ is small. In conclusion, \methodabbr is much more insensitive to the dimensionality turning point than subspace diffusion.

\begin{table*}[t]
\begin{minipage}[c]{0.45\textwidth}
\centering
\caption{\textbf{Synthesis performance} of different models trained on FFHQ $1024\times1024$.}
\label{tab:fid-ffhq1024}
\vspace{-3pt}
\setlength{\tabcolsep}{4pt}
\begin{tabular}{lccc}
\toprule
Model       &\#Params (M)   &NFE    &FID (10k)$\downarrow$ \\
\midrule
Score-SDE   &100            &2000   &52.40 \\
\midrule 
\multirow{3}{*}{CDM}
            &98             &1525   &24.7 \\
            &165            &1525   &17.35 \\
            &286            &1525   &17.24 \\
\midrule
\multirow{2}{*}{DVDP}   &\multirow{2}{*}{105} 
                            &675    &12.43 \\
                            &&1000  &{\bf 10.46} \\
\bottomrule
\end{tabular}
\end{minipage}
\hfill
\begin{minipage}[c]{0.52\textwidth}
\centering
\caption{\textbf{Ablation study} on the number of downsampling times on CelebA 128$\times$128.}
\label{tab:ablation}
\vspace{-3pt}
\setlength{\tabcolsep}{4pt}
\begin{tabular}{lcccc}
\toprule
\makecell[l]{Downsampling \\ times $K$}  &0 &1 &2 &3 \\
\midrule 
FID (50k)                           &6.14   &5.99           &6.10           &6.37 \\
\midrule 
\makecell[l]{Training \\ Speed Up}  &$-$    &1.98$\times$   &2.24$\times$   &2.25$\times$ \\
\midrule
\makecell[l]{Sampling \\ Speed Up}  &$-$    &2.12$\times$   &2.36$\times$   &2.43$\times$ \\
\bottomrule
\end{tabular}
\end{minipage}
\vspace{-5pt}
\end{table*}

\subsection{Ablation Study}\label{subsec:exp-ablation}

We implement ablation study in this section to show that \methodabbr is able to keep effective when the number of downsampling, \textit{i.e.}, $K$, grows. Specifically, we verify that on four different settings of dimensionality turning points $\sT$ for $K=0,1,2,3$. When $K=1$, $\sT$ is set to $\{250\}$. Similarly, when $K=2$ and $K=3$, $\sT$ is set to $\{250, 500\}$ and $\{250, 500, 750\}$, respectively. Furthermore, we use the same noise schedule for these four different settings. \cref{tab:ablation} shows that when the number of downsampling grows, the sampling quality preserves a reasonable level, indicating that \methodabbr can vary dimensionality for multiple times.
\section{Conclusion}\label{sec:conclusion}

This paper generalizes the traditional diffusion process to a dimensionality-varying diffusion process (\methodabbr). The proposed \methodabbr has both theoretical and experimental contributions. Theoretically, we carefully decompose the signal in the diffusion process into multiple orthogonal dynamic attenuation components. With a rigorously deduced approximation strategy, this then leads to a novel reverse process that generates images from much lower dimensional noises compared with the image resolutions. This design allows much faster training and sampling speed of the diffusion models with on-par or even better synthesis performance, and superior performance in synthesizing large-size images of $1024\times1024$ resolution compared with classic methods. The results in this work can promote the understanding and applications of diffusion models in broader scenarios.


{\small
\bibliographystyle{abbrv}
\bibliography{ref}
}

\appendix
\section*{Appendix}
\newcommand{\AppendixPrefix}{A}
\renewcommand\thesection{\Alph{section}}
\setcounter{section}{0}
\renewcommand{\thefigure}{\AppendixPrefix\arabic{figure}}
\setcounter{figure}{0}
\renewcommand{\thetable}{\AppendixPrefix\arabic{table}} 
\setcounter{table}{0}
\renewcommand{\theequation}{\AppendixPrefix\arabic{equation}} 
\setcounter{equation}{0}


\section{Proofs and Derivations}
\label{sec:append-proof-derivation}

\subsection{Details on the Forward Transition Kernel}
\label{subsec:append-forward-transition}

Here we prove that the marginal distributions of each forward \adpabbr given by \cref{eq:xt0-given-x0,eq:xtk-given-x0} can be derived from the forward transition kernel defined by \cref{eq:xtk-given-xt-1}. The proof uses the following basic property of Gaussians 
\begin{equation}
\label{eq:gaussian-property}
\rvz_1\sim \gN(\vmu; \mSigma_1),~\rvz_2 | \rvz_1 \sim \gN(\mA \rvz_1; \mSigma_2) \Rightarrow \rvz_2 \sim \gN(\mA \vmu; \mA \mSigma_1 \mA^\T + \mSigma_2).
\end{equation}

As a prerequisite, we first re-write the Gaussian transition kernel given by \cref{eq:xtk-given-xt-1} as
\begin{equation}
\label{eq:forward-transition}
    \rvx_t^k | \rvx_{t-1}^k \sim \gN(\mU_k \mLambda_{k,t} \mU_k^\T \rvx_{t-1}^k; \mU_k \mL_{k,t}^2 \mU_k^\T),~1\leq t \leq T,~0\leq k \leq K,
\end{equation}
where
\begin{equation}
\label{eq:schedule-relation}
\begin{aligned}
    \mLambda_{k,t} &= \bar{\mLambda}_{k,t-1}^{-1} \bar{\mLambda}_{k,t},\\
    \mL_{k,t} &= (\bar{\mL}_{k,t}^2 - \mLambda_{k,t}^2 \bar{\mL}_{k,t-1}^2 )^{1/2}.
\end{aligned}
\end{equation}
The marginal distributions given by \cref{eq:xt0-given-x0,eq:xtk-given-x0} can also be re-written as
\begin{equation}
\label{eq:forward-marginal}
\rvx_t^k | \rvx_0^k \sim \gN(\mU_k \bar{\mLambda}_{k,t} \mU_k^\T \rvx_0^k; \mU_k \bar{\mL}_{k,t}^2 \mU_k^\T),~1\leq t \leq T,~0\leq k \leq K.
\end{equation}

With \cref{eq:gaussian-property,eq:forward-transition}, we can prove \cref{eq:forward-marginal} by induction:
\begin{enumerate}
    \item For $t=1$, $\rvx_1^k | \rvx_0^k \sim \gN(\mU_k \mLambda_{k,1} \mU_k^\T \rvx_0^k; \mU_k \mL_{k,1}^2 \mU_k^\T)$ is directly defined by \cref{eq:forward-transition}. It satisfies \cref{eq:forward-marginal} since $\mLambda_{k,1} = \bar{\mLambda}_{k,1}$ and $\mL_{k,1} = \bar{\mL}_{k,1}$.

    \item Suppose $\rvx_t^k | \rvx_0^k$ satisfies \cref{eq:forward-marginal}. With the definition of $\rvx_{t+1}^k | \rvx_t^k$ given by \cref{eq:forward-transition} and the property \cref{eq:gaussian-property}, $\rvx_{t+1}^k | \rvx_0^k$ can be derived as
    \begin{equation}
    \label{eq:forward-induction-xt+1}
    \begin{aligned}
    \rvx_{t+1}^k | \rvx_0^k \sim& \gN(\mU_k \mLambda_{k, t+1} \bar{\mLambda}_{k,t} \mU_k^\T \rvx_0^k; \mU_k (\mLambda_{k, t+1}^2 \bar{\mL}_{k,t}^2 + \mL_{k,t+1}^2) \mU_k^\T)\\
    =& \gN(\mU_k \bar{\mLambda}_{k,t+1} \mU_k^\T \rvx_0^k; \mU_k \bar{\mL}_{k,t+1}^2 \mU_k^\T),
    \end{aligned}
    \end{equation}
    where $\bar{\mLambda}_{k,t+1} = \mLambda_{k, t+1} \bar{\mLambda}_{k,t}$ and $\bar{\mL}_{k,t+1}^2 = \mLambda_{k, t+1}^2 \bar{\mL}_{k,t}^2 + \mL_{k,t+1}^2$ can be obtained from \cref{eq:schedule-relation}. Thus, $\rvx_{t+1}^k | \rvx_0^k$ also satisfies \cref{eq:forward-marginal}.
\end{enumerate}
Thus, the proof is completed.

\subsection{Derivation of $q(\rvx_{t-1}^k | \rvx_t^k, \rvx_0^k)$}
\label{subsec:append-forward-post}

Here we derive $q(\rvx_{t-1}^k | \rvx_t^k, \rvx_0^k)$ from the marginal distribution given by \cref{eq:xt0-given-x0,eq:xtk-given-x0} and the forward transition kernel given by \cref{eq:xtk-given-xt-1}. 
By the Bayes' theorem, $q(\rvx_{t-1}^k | \rvx_t^k, \rvx_0^k) \propto q(\rvx_{t-1}^k | \rvx_0^k) q(\rvx_t^k | \rvx_{t-1}^k, \rvx_0^k) = q(\rvx_{t-1}^k | \rvx_0^k) q(\rvx_t^k | \rvx_{t-1}^k)$, where the equality holds because of the Markovian property of $\rvx_0^k \rightarrow \rvx_1^k \cdots \rightarrow \rvx_T^k$. With $q(\rvx_{t-1}^k | \rvx_0^k)$ and $q(\rvx_t^k | \rvx_{t-1}^k)$ given by Eqs. (\textcolor[rgb]{1,0,0}{5}) to (\textcolor[rgb]{1,0,0}{7}), we have
\begin{equation}
\begin{aligned}
    \log q(\rvx_{t-1}^k | \rvx_t^k, \rvx_0^k) =& \log q(\rvx_{t-1}^k | \rvx_0^k) + \log q(\rvx_t^k | \rvx_{t-1}^k) + C_1\\
    =& -\frac{1}{2}(\rvx_{t-1}^k - \mU_k \bar{\mLambda}_{k, t-1} \mU_k^\T \rvx_0^k)^\T \mU_k \bar{\mL}_{k, t-1}^{-2} \mU_k^\T (\rvx_{t-1}^k - \mU_k \bar{\mLambda}_{k, t-1} \mU_k^\T \rvx_0^k)\\
    & -\frac{1}{2}(\rvx_{t}^k - \mU_k \mLambda_{k, t} \mU_k^\T \rvx_{t-1}^k)^\T \mU_k \mL_{k, t}^{-2} \mU_k^\T (\rvx_{t}^k - \mU_k \mLambda_{k, t} \mU_k^\T \rvx_{t-1}^k) + C_2\\
    =& -\frac{1}{2}\big[ {\rvx_{t-1}^k}^\T \mU_k (\bar{\mL}_{k,t-1}^{-2} + \mLambda_{k, t}^2 \mL_{k, t}^{-2}) \mU_k^\T \rvx_{t-1}^k \\
    &-2 (\mU_k \bar{\mLambda}_{k, t-1} \bar{\mL}_{k, t-1}^{-2} \mU_k^\T \rvx_0^k + \mU_k \mLambda_{k, t} \mL_{k, t}^{-2}\mU_k^\T \rvx_t^k)^\T \rvx_{t-1}^k \big] + C_3\\
    =& -\frac{1}{2}\big[ {\rvx_{t-1}^k}^\T \mU_k \mL_{k,t}^{-2} \bar{\mL}_{k,t-1}^{-2} \bar{\mL}_{k,t}^2 \mU_k^\T \rvx_{t-1}^k \\
    &-2 (\mU_k \bar{\mLambda}_{k, t-1} \bar{\mL}_{k, t-1}^{-2} \mU_k^\T \rvx_0^k + \mU_k \mLambda_{k, t} \mL_{k, t}^{-2}\mU_k^\T \rvx_t^k)^\T \rvx_{t-1}^k \big] + C_3\\
    =& -\frac{1}{2} (\rvx_{t-1}^k - \tilde{\vmu}_{k,t})^\T \tilde{\mSigma}_{k, t}^{-2} (\rvx_{t-1}^k - \tilde{\vmu}_{k,t}) + C_4,
\end{aligned}
\end{equation}
where $C_1$, $C_2$, $C_3$ and $C_4$ are constants that do not depend on $\rvx_{t-1}^k$, and
\begin{equation}
\label{eq:forward-post-mean-var}
\begin{aligned}
    \tilde{\vmu}_{k,t} =& \tilde{\vmu}_{k,t}(\rvx_{t}^k, \rvx_0^k) =\mU_k \bar{\mLambda}_{k, t-1} \mL_{k,t}^2 \bar{\mL}_{k,t}^{-2} \mU_k^\T \rvx_0^k + \mU_k \mLambda_{k, t} \bar{\mL}_{k,t-1}^2 \bar{\mL}_{k,t}^{-2} \mU_k^\T \rvx_t^k,\\
    \tilde{\mSigma}_{k,t} =& \mU_k \mL_{k,t}^2 \bar{\mL}_{k, t-1}^2 \bar{\mL}_{k, t}^{-2} \mU_k^\T.
\end{aligned}
\end{equation}
Thus, $q(\rvx_{t-1}^k | \rvx_t^k, \rvx_0^k) = \gN(\rvx_{t-1}^k; \tilde{\vmu}_{k,t}, \tilde{\mSigma}_{k,t})$.

\subsection{Derivation of the Loss Function}
\label{subsec:append-loss}

Here we derive \cref{eq:reverse-loss} from the variational bound on negative log-likelihood
\begin{equation}
\begin{aligned}
    \E_q[-\log p_\theta(\rvx_0^0)] \leq& \E_q\left[-\log \frac{p_\theta(\rvx_{0:T_1}^0,\rvx_{T_1:T_2}^1,\cdots,\rvx_{T_K:T}^K)}{q(\rvx_{1:T_1}^0,\rvx_{T_1:T_2}^1,\cdots,\rvx_{T_K:T}^K | \rvx_0^0)}\right]\\
    =& \E_q\left[ -\log \frac{p_\theta(\rvx_0^0 | \rvx_1^0) p_\theta(\rvx_{1:T_1}^0,\rvx_{T_1:T_2}^1,\cdots,\rvx_{T_K:T-1}^K | \rvx_T^K) p_\theta(\rvx_T^K)}{q(\rvx_{1:T_1}^0,\rvx_{T_1:T_2}^1,\cdots,\rvx_{T_K:T-1}^K | \rvx_T^K, \rvx_0^0) q(\rvx_T^K | \rvx_0^0)}\right] \\
    =& \E_q\Bigg[-\log p_\theta(\rvx_0^0 | \rvx_1^0) -\sum_{k=0}^K \sum_{\substack{t=T_k + 1\\t>1}}^{T_{k+1}}\log \frac{p_\theta(\rvx_{t-1}^k | \rvx_t^k)}{q(\rvx_{t-1}^k | \rvx_t^k, \rvx_0^0)} \\
    &\qquad -\sum_{k=1}^K \log \frac{p_\theta(\rvx_{T_k}^{k-1} | \rvx_{T_k}^k)}{q(\rvx_{T_k}^{k-1} | \rvx_{T_k}^k, \rvx_0^0)} -\log \frac{p_\theta(\rvx_T^K)}{q(\rvx_T^K |\rvx_0^0)} \Bigg]\\
    =& \E_q\Bigg[\underbrace{-\log p_\theta(\rvx_0^0 | \rvx_1^0)}_{L_0} +\sum_{k=0}^K \sum_{\substack{t=T_k + 1\\t>1}}^{T_{k+1}} \underbrace{{\rm D_{KL}}(q(\rvx_{t-1}^k | \rvx_t^k, \rvx_0^k) \Vert p_\theta(\rvx_{t-1}^k | \rvx_t^k))}_{L_{t-1}} \\
    &\qquad +\sum_{k=1}^K \underbrace{{\rm D_{KL}}(q(\rvx_{T_k}^{k-1} | \rvx_{T_k}^k, \rvx_0^{k-1}) \Vert p_\theta(\rvx_{T_k}^{k-1} | \rvx_{T_k}^k))}_{L_k^{\text{down}}} + \underbrace{{\rm D_{KL}}(q(\rvx_T^K |\rvx_0^0) \Vert p_\theta(\rvx_T^K))}_{L_T} \Bigg],
\end{aligned}
\end{equation}
where $L_0, L_T$ and $L_{t-1},~t=2, 3, \cdots, T$ are similar with the definitions in DDPM~\cite{ho2020denoising}, and $L_{k}^\text{down}$ is a new term and can be viewed as the loss at the dimensionality turning point $T_k$. As defined in \cref{eq:reverse-up}, $p_\theta(\rvx_{T_k}^{k-1} | \rvx_{T_k}^k))$ has no learnable parameters, so we do not optimize $L_{k}^\text{down}$.

As for $L_{t-1}$, it is the KL divergence of two Gaussians and can be calculated as
\begin{equation}
    L_{t-1} = \E_q \bigg[ \frac{1}{2}\Vert \mSigma_t^{-1/2} \mU_k^\T (\tilde{\vmu}_{k,t}(\rvx_t^k, \rvx_0^k) - \vmu_\theta(\rvx_t^k, t))\Vert^2 \bigg] + C,
\end{equation}
where $C$ is a constant that does not depend on $\theta$, $k$ satisfies $T_k < t \leq T_{k+1}$, $\tilde{\vmu}_{k,t}(\rvx_t^k, \rvx_0^k)$ is the mean of $q(\rvx_{t-1}^k | \rvx_t^k, \rvx_0^k)$ given by \cref{eq:forward-post-mean-var}, and $\vmu_\theta$ is the mean of $p_\theta(\rvx_{t-1}^k | \rvx_t^k)$ given by \cref{eq:reverse-mean}.

With \cref{eq:xt0-given-x0,eq:xtk-given-x0}, $L_{t-1}$ can be represented by reparameterization trick as
\begin{equation}
\begin{aligned}
    L_{t-1} =& \E_{\rvx_0^k,\rvepsilon^k} \bigg[ \frac{1}{2}\big\Vert \mSigma_t^{-1/2} \mU_k^\T (\tilde{\vmu}_{k,t}(\rvx_t^k(\rvx_0^k, \rvepsilon^k), \mU_k \bar{\mLambda}_{k,t}^{-1} \mU_k^\T \rvx_t^k(\rvx_0^k, \rvepsilon^k) - \mU_k \bar{\mLambda}_{k,t}^{-1} \bar{\mL}_{k,t} \mU_k^\T \veps^k)\\
    & -\vmu_\theta(\rvx_t^k(\rvx_0^k, \rvepsilon^k), t))\big\Vert^2 \bigg] + C\\
    =& \E_{\rvx_0^k,\rvepsilon^k} \bigg[ \big\Vert \mW_t (\rvepsilon^k - \rvepsilon_\theta(\rvx_{t}^k(\rvx_0^k, \rvepsilon^k), t)) \big\Vert^2 \bigg] + C,
\end{aligned}
\end{equation}
where the final equality is obtained by plugging \cref{eq:reverse-mean} and \cref{eq:forward-post-mean-var} into it, and $\mW_t = \frac{1}{\sqrt{2}}\mSigma_t^{-1/2} \mLambda_{k,t}^{-1} \mL_{k,t}^2 \bar{\mL}_{k,t}^{-1} \mU_k^\T$.

Finally, by setting $\mW_t = \mI$ as in DDPM~\cite{ho2020denoising}, we can obtain \cref{eq:reverse-loss}.

\subsection{Proof of \Cref{prop:xt-diff}}
\label{subsec:append-proof-prop1}

\indent According to the inequality between JSD and \textit{total variation}, we have
\begin{equation}
\label{eq:prop1-JSD-total-variation}
{\rm JSD}(p_1 || p_2) \leq \frac{1}{2} \int |p_1(x) - p_2(x)| d\vx.
\end{equation}

The RHS (right-hand side) of \cref{eq:prop1-JSD-total-variation} satisfies
\begin{equation}
\begin{aligned}
\label{eq:prop1-total-variation}
\frac{1}{2} \int |p_1(\vx) - p_2(\vx)| d\vx =& \frac{1}{2} \int \left|\E_{\rvx_0 \sim p}[ p_1(\vx | \rvx_0) - p_2(\vx | \rvx_0)]\right| d\vx \\
\leq& \frac{1}{2} \int \E_{\rvx_0 \sim p}
\left[ \left| p_1(\vx | \rvx_0) - p_2(\vx | \rvx_0) \right| \right] d\vx\\
=& \frac{1}{2} C_1 \int \E_{\rvx_0 \sim p}
\bigg[ \Big|  \exp\big(-\frac{1}{2} (\vx - \mA_1 \rvx_0)^\T \Sigma^{-1} (\vx - \mA_1 \rvx_0)\big) \\
&- \exp\big(-\frac{1}{2} (\vx - \mA_2 \rvx_0)^\T \Sigma^{-1} (\vx - \mA_2 \rvx_0)\big) \Big| \bigg] d\vx,
\end{aligned}
\end{equation}
where $C_1 = (2\pi)^{-1/2} {\rm det}(\mSigma)^{-1/2}$.

According to the mean value theorem, for each $\rvx_0$ and $\vx$, there exists $\theta=\theta(\rvx_0, \vx)\in [0,1]$ such that $\vxi = \theta(\vx - \mA_1 \rvx_0) + (1 - \theta)(\vx - \mA_2 \rvx_0) = \vx - [\theta \mA_1 + (1 - \theta)\mA_2]\rvx_0$ satisfies
\begin{equation}
\begin{aligned}
\label{eq:prop1-p1-p2-diff}
&\exp\big(-\frac{1}{2} (\vx - \mA_1 \rvx_0)^\T \Sigma^{-1} (\vx - \mA_1 \rvx_0)\big)
- \exp\big(-\frac{1}{2} (\vx - \mA_2 \rvx_0)^\T \Sigma^{-1} (\vx - \mA_2 \rvx_0)\big)\\
=& \vxi^\T \mSigma^{-1} (\mA_1 - \mA_2)\rvx_0 \exp\big(-\frac{1}{2}\vxi^T \mSigma^{-1} \vxi\big) \\
=& F \cdot \exp\big(-\frac{1}{4}\vxi^T \mSigma^{-1} \vxi\big),
\end{aligned}
\end{equation}
where $F =  \vxi^\T \mSigma^{-1} (\mA_1 - \mA_2)\rvx_0 \exp\big(-\frac{1}{4}\vxi^T \mSigma^{-1} \vxi\big)$, and $|F|$ satisfies the following inequality
\begin{equation}
\begin{aligned}
\label{eq:prop1-F}
|F| =& \Big|\frac{(\mSigma^{1/2} \vxi)^\T}{\Vert \mSigma^{-1/2} \vxi \Vert_2} \mSigma^{-1/2} (\mA_1 - \mA_2)\rvx_0 \Big| \cdot \Vert \mSigma^{-1/2} \vxi \Vert_2 \exp\big(-\frac{1}{4} \Vert \mSigma^{-1/2} \vxi \Vert_2^2\big)\\
\leq& C_2 \Vert \mSigma^{-1/2} (\mA_1 - \mA_2)\rvx_0 \Vert_2\\
\leq& C_2 B \Vert \mSigma^{-1/2} (\mA_1 - \mA_2) \Vert_2, 
\end{aligned}
\end{equation}
where $C_2 = \max_{a \geq 0} a e^{-\frac{1}{4} a^2} = \sqrt{2}e^{-\frac{1}{2}}$, and $B$ is the upper bound of $\Vert \rvx_0 \Vert_2$ as assumption. 

Combining \cref{eq:prop1-total-variation,eq:prop1-p1-p2-diff,eq:prop1-F}, we have
\begin{equation}
\begin{aligned}
\label{eq:prop1-total-variation-2}
    \frac{1}{2} \int |p_1(\vx) - p_2(\vx)| d\vx \leq& \frac{1}{2} C_1 C_2 B \Vert \mSigma^{-1/2} (\mA_1 - \mA_2) \Vert_2 \int \E_{\rvx_0 \sim p}\Big[\exp\big(-\frac{1}{4}\vxi^\T \mSigma^{-1} \vxi\big)\Big] d\vx,
\end{aligned}
\end{equation}
where $\vxi = \vx - [\theta \mA_1 + (1 - \theta)\mA_2]\rvx_0$. Now we only need to prove that the RHS of \cref{eq:prop1-total-variation-2} $\leq$ the LHS (left-hand side) of \cref{eq:prop-error-bound}. 

Let $\rvz = \mSigma^{-1/2} [\theta \mA_1 + (1 - \theta)\mA_2]\rvx_0$, then $\vxi^\T \mSigma^{-1} \vxi = \Vert \mSigma^{-1/2} \vx - \rvz\Vert_2^2$, and $\rvz$ satisfies
\begin{equation}
\label{eq:prop1-z}
\begin{aligned}
    \Vert \rvz \Vert_2 =& \Vert \mSigma^{-1/2} [\theta \mA_1 + (1 - \theta)\mA_2]\rvx_0 \Vert_2\\
    \leq& B \Vert \mSigma^{-1/2} [\theta \mA_1 + (1 - \theta)\mA_2] \Vert_2\\
    \leq& B \Vert \mSigma^{-1/2} \mA_1 \Vert_2,
\end{aligned}
\end{equation}
where the last inequality is derived from the assumption that $\mA_1 \succeq \mA_2 \succeq \vzero$.

Let $\sD = \{ \vx : \Vert \mSigma^{-1/2} \vx \Vert_2 \leq r\}$, where $r = 2 B \Vert \mSigma^{-1/2} \mA_1 \Vert_2$, thus $\Vert \rvz \Vert_2 \leq \frac{1}{2} r$ according to \cref{eq:prop1-z}. Then the integration in \cref{eq:prop1-total-variation-2} can be split into two regions as
\begin{equation}
\begin{aligned}
\label{eq:prop1-int}
    \int \E_{\rvx_0 \sim p}\Big[\exp\big(-\frac{1}{4}\vxi^\T \mSigma^{-1} \vxi\big)\Big] d\vx =& \int_\sD \E_{\rvx_0 \sim p}\Big[\exp\big(-\frac{1}{4}\Vert \mSigma^{-1/2}\vx - \rvz \Vert_2^2\big)\Big] d\vx\\
    & + \int_{\sD^{\rm C}} \E_{\rvx_0 \sim p}\Big[\exp\big(-\frac{1}{4}\Vert \mSigma^{-1/2}\vx - \rvz \Vert_2^2\big)\Big] d\vx\\
    \leq& \int_\sD 1 d\vx + \int \exp\big(-\frac{1}{16}\Vert \mSigma^{-1/2}\vx \Vert_2^2\big) d\vx\\
    \leq& V_d(r) {\rm det}(\Sigma)^{1/2} + 2\sqrt{2} (2\pi)^{d/2} {\rm det}(\Sigma)^{1/2},
\end{aligned}
\end{equation}
where $V_d(\cdot)$ is the volume of $d$-dimensional sphere with respect to the radius.

Combining \cref{eq:prop1-JSD-total-variation,eq:prop1-total-variation-2,eq:prop1-int}, we can get \Cref{prop:xt-diff}.

\subsection{Proof of \Cref{thm:reverse-error}}
\label{subsec:append-proof-thm1}

We first prove that $q(\rvx_{T_k}^{k-1})$ and $p(\rvx_{T_k}^{k-1})$ defined in \Cref{thm:reverse-error} satisfy the conditions claimed in \Cref{prop:xt-diff}.

$q(\rvx_{T_k}^k)$ is the marginal distribution of $q(\rvx_0^k)q(\rvx_{T_k}^k | \rvx_0^k)$ where $q(\rvx_{T_k}^k | \rvx_0^k)$ is defined by \cref{eq:xt0-given-x0,eq:xtk-given-x0}. $q(\rvx_{T_k}^k | \rvx_0^k)$ can also be expressed as
\begin{equation}
\label{eq:append-q-k}
    q(\rvx_{T_k}^k | \rvx_0^k) = \gN(\rvx_{T_k}^k; \mU_k \bar{\mLambda}_{k,T_k} \mU_k^\T \rvx_0^k, \mU_k \bar{\mL}_{k,T_k}^2 \mU_k^\T).
\end{equation}

Similarly, $q(\rvx_{T_k}^{k-1})$ is the marginal distribution of $q(\rvx_0^{k-1})q(\rvx_{T_k}^{k-1} | \rvx_0^{k-1})$ where $q(\rvx_{T_k}^{k-1} | \rvx_0^{k-1})$ can be expressed as
\begin{equation}
\label{eq:append-q-k-1}
    q(\rvx_{T_k}^{k-1} | \rvx_0^{k-1}) = \gN(\rvx_{T_k}^{k-1}; \mU_{k-1} \bar{\mLambda}_{{k-1},T_k} \mU_{k-1}^\T \rvx_0^{k-1}, \mU_{k-1} \bar{\mL}_{{k-1},T_k}^2 \mU_{k-1}^\T).
\end{equation}

By definition, $p(\rvx_{T_k}^{k-1})$ is the marginal distribution of $q(\rvx_{T_k}^k)p(\rvx_{T_k}^{k-1} | \rvx_{T_k}^k)$, where $p(\rvx_{T_k}^{k-1} | \rvx_{T_k}^k)$ is defined by \cref{eq:reverse-up} and can be expressed as
\begin{equation}
\label{eq:append-p-trans}
    p(\rvx_{T_k}^{k-1} | \rvx_{T_k}^k) = \gN(\rvx_{T_k}^{k-1}; \gD_k^\T \rvx_{T_k}^k, \mU_{k-1} \Delta \mL_{k-1}^2 \mU_{k-1}^\T).
\end{equation}

To transform $p(\rvx_{T_k}^{k-1})$ into the form in \Cref{prop:xt-diff}, we construct a Markov chain $\rvx_0^{k-1} \rightarrow \rvx_{T_k}^k \rightarrow \rvx_{T_k}^{k-1}$, where $\rvx_0^{k-1} \sim q(\rvx_0^{k-1})$, $\rvx_{T_k}^k | \rvx_0^{k-1} \sim q(\rvx_{T_k}^k | \rvx_0^{k-1}) = q(\rvx_{T_k}^k | \gD_k \rvx_0^{k-1})$ and $\rvx_{T_k}^{k-1} | \rvx_{T_k}^k \sim p(\rvx_{T_k}^{k-1} | \rvx_{T_k}^k)$.
Thus $p(\rvx_{T_k}^{k-1})$ is also the marginal distribution of the joint distribution defined by the Markov chain.
This joint distribution can be factorized as $q(\rvx_0^{k-1}) p_q(\rvx_{T_k}^{k-1} | \rvx_0^{k-1})$, where $p_q(\rvx_{T_k}^{k-1} | \rvx_0^{k-1})$ is the marginal distribution of $q(\rvx_{T_k}^k | \rvx_0^{k-1}) p(\rvx_{T_k}^{k-1} | \rvx_{T_k}^k)$, and can be derived from \cref{eq:append-q-k,eq:append-p-trans} by using \cref{eq:gaussian-property}
\begin{equation}
\label{eq:append-pq}
\begin{aligned}
    p_q(\rvx_{T_k}^{k-1} | \rvx_0^{k-1}) =& \gN(\rvx_{T_k}^{k-1}; \gD_k^\T \mU_k \bar{\mLambda}_{k,T_k} \mU_k^\T \gD_k \rvx_0^{k-1}, \mU_{k-1} \Delta \mL_{k-1}^2 \mU_{k-1}^\T + \gD_k^\T \mU_k \bar{\mL}_{k,T_k}^2 \mU_k^\T \gD_k)\\
    =& \gN(\rvx_{T_k}^{k-1}; \mU_{k-1} (\bar{\mLambda}_{k-1,T_k} - \Delta \mLambda_{k-1}) \mU_{k-1}^\T \rvx_0^{k-1}, \mU_{k-1} \mL_{k-1, T_k}^2 \mU_{k-1}^\T),
\end{aligned}
\end{equation}
where $\Delta \mLambda_{k-1} = {\rm diag}(\bar{\lambda}_{k-1, T_k}\mI_{d_{k-1}}, \mO_{\bar{d}_k})$.

Thus, $q(\rvx_{T_k}^{k-1} | \rvx_0^{k-1})$ given by \cref{eq:append-q-k-1} and $p_q(\rvx_{T_k}^{k-1} | \rvx_0^{k-1})$ given by \cref{eq:append-pq} satisfy conditions of $p_1$ and $p_2$ claimed in \Cref{prop:xt-diff} respectively. And $\Vert \rvx_0^{k-1} \Vert$ satisfies
\begin{equation}
    \Vert \rvx_0^{k-1} \Vert = \Vert \overline{\gD}_{k-1} \rvx_0^0 \Vert \leq \Vert \rvx_0^0 \Vert \leq \sqrt{d}.
\end{equation}

Finally, substituting all corresponding variables into \cref{eq:prop-error-bound}, we can obtain \cref{eq:thm1-error-bound}.

\section{Implementation Details}
\label{sec:append-implement}

In this section, we will give more details on the implementation of our \methodabbr. 
\cref{alg:train,alg:sample} display the complete training and sampling procedures respectively.

\noindent
\begin{minipage}{0.47\linewidth}
\begin{algorithm}[H]
    \caption{Training}\label{alg:train}
    \begin{algorithmic}[1]
    \Repeat
    \State Sample $k$ from $(\{0,\cdots,K\})$
    \State $t \sim {\rm Uniform}(\{T_k+1, \cdots,T_{k+1}\})$
    \State $\rvx_0^0 \sim q(\rvx_0^0)$
    \State $\rvepsilon^k \sim \gN({\bm 0}; \mI_{\bar{d}_k})$
    \State $\rvx_0^k \gets \overline{\gD}_k \rvx_0^0$
    \State $\rvx_t^k \gets \mU_k \bar{\mLambda}_{k,t} \mU_k^\T \rvx_0^k + \mU_k \bar{\mL}_{k,t} \mU_k^\T \rvepsilon^k$
    \State Take gradient descent step on
    \Statex \qquad \quad $\nabla_\theta \Vert \rvepsilon^k - \veps_\theta (\rvx_t^k, t) \Vert^2$
    \Until{converged}
    \end{algorithmic}
\end{algorithm}
\end{minipage}
\hfill
\begin{minipage}{0.53\linewidth}
\begin{algorithm}[H]
    \caption{Sampling}\label{alg:sample}
    \begin{algorithmic}[1]
    \State $\rvx_{T}^K \sim \gN({\bm 0}; \mI_{\bar{d}_K})$
    \For{$k = K, \cdots, 0$}
        \For{$t = T_{k+1}, \cdots, T_k + 1$}
            \State $\rvepsilon^k \sim \gN({\bm 0}; \mI_{\bar{d}_k})$
            \State $\rvx_{t-1}^k \gets \mU_k \bar{\mLambda}_{k,t}^{-1} (\mU_k^\T \rvx_t^k - \bar{\mL}_{k,t} \mU_k^\T \veps_\theta(\rvx_t^k, t)) + \mSigma_t \rvepsilon^k $
        \EndFor
        \If{$k > 0$}
            \State $\rvepsilon^{k-1} \sim \gN({\bm 0}; \mI_{\bar{d}_{k-1}})$
            \State $\rvx_{T_k}^{k-1} \gets \gD_k^\T \rvx_{T_k}^{k} + \mU_{k-1} \Delta\mL_{k-1} \mU_{k-1}^\T \rvepsilon^{k-1}$
        \EndIf
    \EndFor
    \State \Return $\rvx_0^0$
    \end{algorithmic}
\end{algorithm}
\end{minipage}

\subsection{Choice of Downsampling Operator $\gD_k$}
Since an image pixel is usually similar with its neighbours, we can simply choose $\gD_k$ to be a $2\times 2$ average-pooling operator for each $k=1,\cdots,K$ as in subspace diffusion~\cite{jing2022subspace} to maintain the main component of an image. Under this choice, the dimensionality will be reduced from $\bar{d}_{k-1}$ to $\bar{d}_k = \frac{1}{4}\bar{d}_{k-1}$ after each downsampling operation, as mentioned in \cref{subsec:exp-setup}. 

The above choice needs a simple modification, multiplication by 2 after the average-pooling operation, to ensure that the matrix $\gD_k$ satisfies
\begin{equation}
\label{eq:Dk-condition}
\begin{aligned}
    \gD_k [\mN_{k-1}, \mB_{k-1}] = [\bm{0}, \mU_k],
\end{aligned}
\end{equation}
where $\mN_{k-1} \in \sR^{\bar{d}_{k-1}\times d_{k-1}}$ and $\mB_{k-1} \in \sR^{\bar{d}_{k-1} \times \bar{d}_k}$ satisfies $\mU_{k-1} = [\mN_{k-1}, \mB_{k-1}]$. Under this condition, the matrix $\gD_k \in \sR^{\bar{d}_k\times \bar{d}_{}k-1}$ is row-orthogonal, and $\gD_k^\T$ (\textit{i.e.}, the transpose of $\gD_k$) is just the corresponding upsampling operator.

\subsection{Attenuation Coefficient $\bar{\lambda}_{k,t}$}
As mentioned in \cref{subsec:method-forward}, only $\bar{\lambda}_{k,t}$ is required to be approximate zero at dimensionality turning point $T_{k+1}$ for $k=0,\cdots,K-1$, thus we only need to decrease $\bar{\lambda}_{k,t}$ when $T_k < t \leq T_{k+1}$ and keep $\bar{\lambda}_{i,t},~i\neq k$ unchanged. In experiment, we decrease $\bar{\lambda}_{k,t}$ in an exponential form. Thus, for $\bar{\lambda}_{k,t},~k=0,1,\cdots,K-1$, they are set in the following manner:
\begin{equation}
    \bar{\lambda}_{k,t} = 
    \begin{cases}
        1, & t \leq T_k\\
        \bar{\lambda}_{\min}, & t > T_{k+1}\\
        \bar{\lambda}_{\min}^{(t - T_k)/(T_{k+1} - T_k)}, & T_{k} < t \leq T_{k+1},
    \end{cases}
\end{equation}
where $\bar{\lambda}_{\min} \in (0, 1)$ is a shared hyperparameters for $\bar{\lambda}_{0,t},\bar{\lambda}_{1,t},\cdots,\bar{\lambda}_{K-1,t}$. For all experiments, we set $\bar{\lambda}_{\min} = 0.01$. As for $k=K$, we set $\bar{\lambda}_{K, t} = 1$ for all $t$. This schedule means that between two adjacent dimensionality turning points $T_{k}$ and $T_{k+1}$, we only attenuate one data component $\rvv_k^k$. Once we set $\bar{\lambda}_{k,t}$ for each $k$ and $t$, hyperparameters ${\lambda}_{k,t},~\bar{\mLambda}_{k,t},~\mLambda_{k,t}$ are determined.

\subsection{Noise Schedule $\bar{\sigma}_{k,t}$}
\label{subsec:append-noise-schedule}

At each timestep $t$, we set $\sigma_{0,t} = \sigma_{1,t} = \cdots, =\sigma_{K,t} \defeq \sigma_t$, which means that the added noise at each step is symmetric, similar with that in DDPM~\cite{ho2020denoising}.
Rather than setting $\sigma_t$ directly, we first determine $\bar{\sigma}_t = \sum_{s=1}^t \sigma_s$, then obtain $\sigma_t$ by $\sigma_t = \bar{\sigma}_t / \bar{\sigma}_{t-1}$.

Since we choose subspaces in which the main components of images stay, the image signal will not lose much components when getting close to the subspace and downsampled to a smaller size. However, Gaussian noise does not have this property and can lose large parts of components in the downsampling operation. Thus, the signal-to-noise (SNR) ratio at the last timestep $T$ will be smaller than that in DDPM~\cite{ho2020denoising} if we just use the same noise schedule. Suppose at $T_k,~k=1,2,\cdots,K$, $\dxkt{k-1}{T_k} \in \sR^{\bar{d}_{k-1}}$ is downsampled to $\dxkt{k}{T_k} \in \sR^{\bar{d}_{k}}$ with downsamling factor $f_k = \bar{d}_{k-1} / \bar{d}_k$, then the noise shedule is adapted as \Cref{alg:noise-schedule}, which can approximately keep the SNR at the last timestep meanwhile maintaining the continuity of $\bar{\sigma}$.
\begin{algorithm}
\caption{Adaptation on Noise Schedule}\label{alg:noise-schedule}
\begin{algorithmic}[1]
\State Initialize $\bar{\alpha}_{0:T}$ as in DDPM
\State $\bar{\sigma} \gets \sqrt{\frac{1}{\bar{\alpha}} - 1}$
\For{$k = 1, \cdots, K$}
\State $\bar{\sigma}_{T_k:T} \gets \bar{\sigma}_{T_k - 1} + f_k \cdot (\bar{\sigma}_{T_k:T} - \bar{\sigma}_{T_k - 1})$
\EndFor
\end{algorithmic}
\end{algorithm}

\subsection{Simplification of Matrix Multiplication $\mU_k \mG_k \mU_k^\T$}
With the above choices of attenuation coefficients and noise schedule, all matrix multiplications with the form of $\mU_k \mG_k \mU_k^\T$ in the implementation of \methodabbr can be expressed by downsampling operator $\gD_{k+1}$ and upsampling operator $\gD_{k+1}^\T$, since each diagonal matrix $\mG_k$ only includes two different elements and can be expressed in the form of $\mG = {\rm diag}(a_k\mI_{d_k}, b_k\mI_{\bar{d}_{k+1}})$. Thus, $\mU_k \mG_k \mU_k^\T$ can be expressed as
\begin{equation}
\begin{aligned}
    \mU_k \mG_k \mU_k^\T =& [\mN_{k}, \mB_{k]}]
    \begin{bmatrix}
        a_k\mI_{d_k} &{\bm 0}\\
        {\bm 0} &b_k\mI_{\bar{d}_{k+1}}
    \end{bmatrix}
    [\mN_{k}, \mB_{k]}]^\T\\
    =& a_k \mI_{\bar{d}_k} + [\mN_{k}, \mB_{k]}]
    \begin{bmatrix}
        {\bm 0} &{\bm 0}\\
        {\bm 0} &(b_k - a_k)\mI_{\bar{d}_{k+1}}
    \end{bmatrix}
    [\mN_{k}, \mB_{k]}]^\T\\
    =& a_k \mI_{\bar{d}_k} + (b_k - a_k) \mB_k \mB_k^\T\\
    =& a_k \mI_{\bar{d}_k} + (b_k - a_k) \gD_{k+1}^\T \gD_{k+1},
\end{aligned}
\end{equation}
where the last equality can be derived from \cref{eq:Dk-condition}.

\section{Experiments on DDIM Sampling}
\label{sec:append-exp}

To demonstrate that our \methodabbr is compatible with DDIM~\cite{song2020denoising}, an accelerated sampling method, we apply DDIM to our models trained on LSUN Church $256\times256$ and FFQH $256\times 256$. The results are shown in \cref{tab:ddim}. In experiment, we find that it is beneficial for \methodabbr to add noises in some middle steps of sampling, unlike DDIM that sets all inserted noises to zeros. Specifically, we set $\eta_t = 1$ for $t = T_1 - \lfloor T_1 /4 \rfloor, \cdots, T_1 + \lceil (T_2 - T_1) / 2 \rceil$ and $\eta_t = 0$ otherwise, where $\eta_t \in [0, 1]$ controls the strength of added noise as in DDIM~\cite{song2020denoising} for timestep $t$. This adaption is marked by $^\ast$ in \cref{tab:ddim}. 
\begin{table}[h]
\caption{
    \textbf{Quantitative comparison} measured in FID. DDIM$^\ast$ denotes an adapted DDIM sampling method.
}
\label{tab:ddim}
\vspace{-15pt}
\begin{center}
\setlength{\tabcolsep}{4pt}
\begin{tabular}{l|ccc|ccc}
\toprule
Dataset &\multicolumn{3}{c|}{Church $256\times 256$} &\multicolumn{3}{c}{FFHQ $256\times 256$}\\
\#Steps                 &50         &100        &200            &50             &100        &200\\
\midrule 
DDIM Baseline           &10.44      &10.22      &10.26          &12.32          &10.80      &10.19\\
DDIM$^\ast$ Baseline    &9.36       &8.91       &9.03           &13.33          &10.28      &9.06\\
DDIM$^\ast$ DVDP        &{\bf 8.52} &{\bf 7.33} &{\bf 7.32}    &{\bf 12.01}    &{\bf 8.39} &{\bf 7.04}\\
\bottomrule
\end{tabular}
\end{center}
\vspace{-25pt}
\end{table}

\end{document}